\documentclass[letterpaper]{article} % DO NOT CHANGE THIS
\usepackage{aaai24}  % DO NOT CHANGE THIS
\usepackage{times}  % DO NOT CHANGE THIS
\usepackage{helvet}  % DO NOT CHANGE THIS
\usepackage{courier}  % DO NOT CHANGE THIS
\usepackage[hyphens]{url}  % DO NOT CHANGE THIS
\usepackage{graphicx} % DO NOT CHANGE THIS
\usepackage{natbib}  % DO NOT CHANGE THIS AND DO NOT ADD ANY OPTIONS TO IT
\usepackage{caption} % DO NOT CHANGE THIS AND DO NOT ADD ANY OPTIONS TO IT
\usepackage{algorithm}
\usepackage{algorithmic}

\usepackage{newfloat}
\usepackage{listings}
\DeclareCaptionStyle{ruled}{labelfont=normalfont,labelsep=colon,strut=off} % DO NOT CHANGE THIS
\floatstyle{ruled}
\newfloat{listing}{tb}{lst}{}
\floatname{listing}{Algorithm}
\nocopyright % -- Your paper will not be published if you use this command
\usepackage{booktabs}
\usepackage{multirow}

\usepackage{amsmath}
\usepackage{amsthm}

\DeclareMathOperator*{\argmax}{arg\,max}

\DeclareMathOperator*{\argtopk}{arg\,topk}

\DeclareMathOperator*{\kthmax}{k-th\,max}

\usepackage[english]{babel}

\newtheorem{proposition}{Proposition}

\newcommand{\BOS}{\textsc{BOS}}
\newcommand{\EOS}{\textsc{EOS}}
\newcommand{\vx}{\mathbf{x}}
\newcommand{\vy}{\mathbf{y}}
\newcommand{\vocab}{\mathcal{V}}

\newcommand{\nmax}{n_{\mathrm{max}}}

\newcommand{\Analyzed}{{Lookahead Beam Search}}
\newcommand{\analyzed}{{lookahead beam search}}
\newcommand{\Proposed}{{Lookbehind Heuristic Beam Search}}
\newcommand{\proposed}{{lookbehind heuristic beam search}}
\newcommand{\ana}{{LBS}}
\newcommand{\pro}{{LHBS}}

\title{On the Depth between Beam Search and Exhaustive Search for Text Generation}
\author{
    Yuu Jinnai, Tetsuro Morimura, Ukyo Honda
}
\affiliations{
    CyberAgent, Inc.
    \{jinnai\_yu,morimura\_tetsuro,honda\_ukyo\}@cyberagent.co.jp
}

\begin{document}

% TODO: REMOVE THIS BEFORE SUBMISSION
% \SetBgContents{UNDER REVIEW; DO NOT CIRCULATE; YJ, TM, UH}
% \SetBgScale{2}
% \SetBgAngle{0}
% \SetBgOpacity{0.6}
% \SetBgColor{red}
% \SetBgPosition{current page.south west}
% \SetBgHshift{5cm}
% \SetBgVshift{13.5cm}

\maketitle

\begin{abstract}
% \todo{title of the figures}

% \todo{restructure Sec 3.2}
% \todo{replace results of Sec 4 with 500 sentences}
% The performance of natural language generation systems has improved substantially with neural networks. At test time they typically employ beam search to find optimal predictions. However, due to model errors, a large beam size can lead to deteriorating performance according to the evaluation metric. For this reason, it is common to rerank the output of beam search,
% Previous work has examined the effect of beam width and has reported that increasing the width beyond a certain point decreases performance, which is known as the ``curse'' of beam search.
% In this work, we investigate between the two ends in terms of search depth, which has been less explored in previous work.

% Beam Search
Beam search and exhaustive search are two extreme ends of text decoding algorithms with respect to the search depth. 
Beam search is limited in both search width and depth, whereas exhaustive search is a global search that has no such limitations. 
Surprisingly, beam search is not only computationally cheaper but also performs better than exhaustive search despite its higher search error. 
Plenty of research has investigated a range of beam widths, from small to large, and reported that a beam width that is neither too large nor too small is desirable.
However, in terms of search depth, only the two extreme ends, beam search and exhaustive search are studied intensively.
In this paper, we examine a range of search depths between the two extremes to discover the desirable search depth.
To this end, we introduce Lookahead Beam Search (LBS), a multi-step lookahead search that optimizes the objective considering a fixed number of future steps.
Beam search and exhaustive search are special cases of LBS where the lookahead depth is set to $0$ and $\infty$, respectively.
We empirically evaluate the performance of LBS and find that it outperforms beam search overall on machine translation tasks. 
The result suggests there is room for improvement in beam search by searching deeper.
Inspired by the analysis, we propose Lookbehind Heuristic Beam Search, a computationally feasible search algorithm that heuristically simulates LBS with 1-step lookahead.
The empirical results show that the proposed method outperforms vanilla beam search on machine translation and text summarization tasks.

\end{abstract}

\section{Introduction}

% \begin{figure}[t]
%     \centering
%     \includegraphics[width=\columnwidth]{depth-bleu.pdf}
%     % \includegraphics[width=\columnwidth]{depth-bleu.png}
%     \caption{BLEU Improvement over vanilla beam search. Average over beam widths ($k=5$, $10$, $15$, and $20$) are plotted in bold. The shaded area shows an unbiased standard error of the mean. Evaluated on first 100 sentences of WMT'14 En-Fr dataset.}
%     \label{fig:locality}
% \end{figure}

\begin{figure}[t]
    \centering
    \includegraphics[width=\columnwidth]{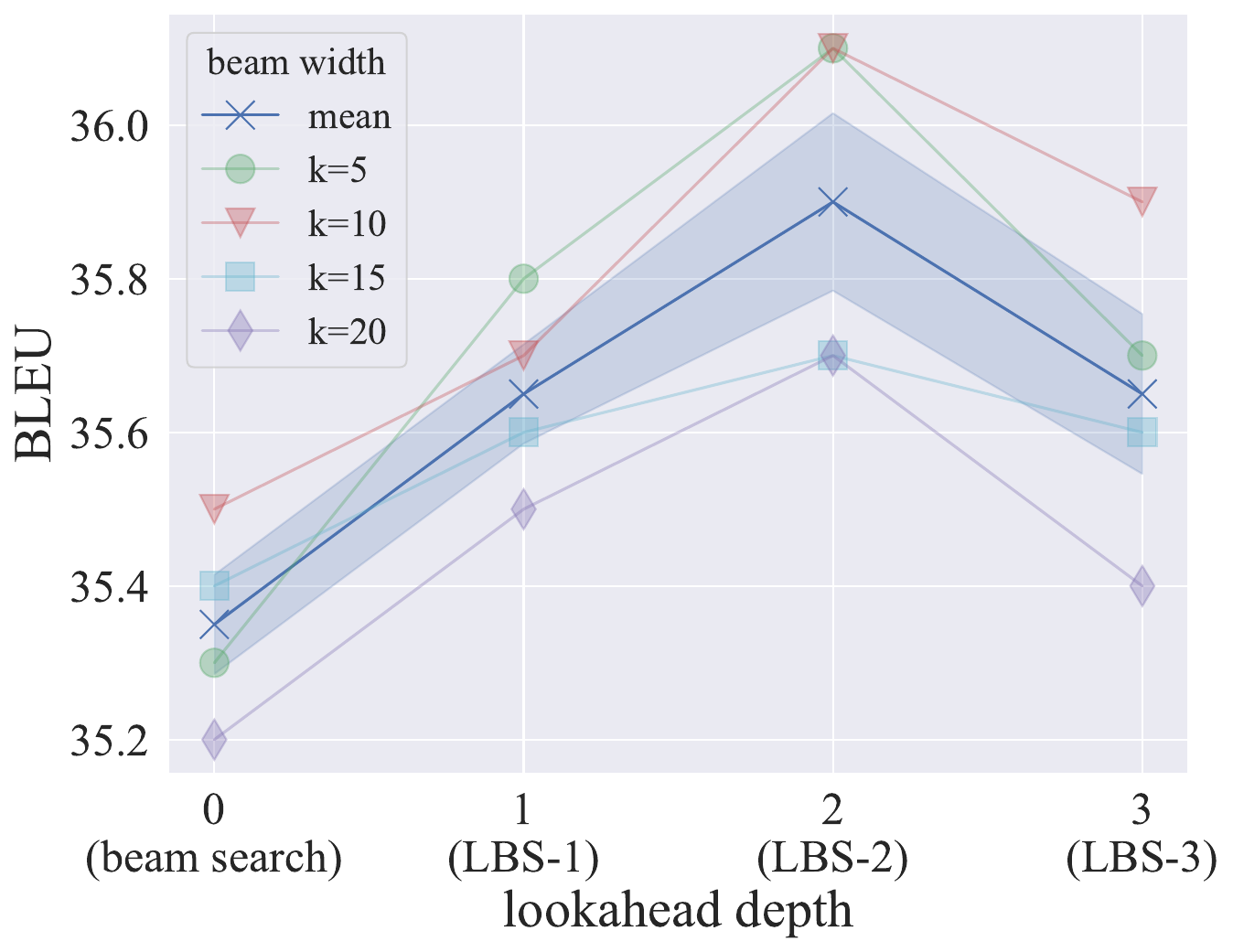}
    \caption{Results on machine translation by beam search and \analyzed{} (LBS) with lookahead depth 1, 2, and 3. The bold line represents the mean and the shaded area shows the standard error. Evaluated on the first 100 sentences of WMT'14 En-Fr dataset.}
    \label{fig:locality}
\end{figure}

%%%%%%%%%%%%%%%%%%%%%%%%%%%%%%%%%%%%%%
% Natural Language Generation Task
The goal of natural language generation is to generate text representing structured information that is both fluent and contains the appropriate information.
% Modern probabilistic models have shown their efficiency at modeling natural language in many different tasks \cite{}. However, 
% Beam search is used for directed generation
One of the key design decisions in text generation is the choice of decoding strategy. The decoding strategy is the decision rule used to generate strings from a probabilistic model (e.g., Transformer; \citealp{NIPS2017_3f5ee243}).

%%%%%%%%%%%%%%%%%%%%%%%%%%%%%%%%%%%%%%
% Exhaustive search
A straightforward solution is to \textbf{exhaustively search} for the strings with the highest probability with respect to the model. This is known as \textbf{maximum a posteriori (MAP) decoding}. Not only exhaustive search is computationally infeasible, but surprisingly, it is known to produce low-quality text \cite{murray-chiang-2018-correcting,pmlr-v97-cohen19a}. For example, \citet{stahlberg-byrne-2019-nmt} reports that in machine translation tasks, the highest-probability string is often the empty string.

%%%%%%%%%%%%%%%%%%%%%%%%%%%%%%%%%%%%%%
% Beam search 
\textbf{Beam search} has been the go-to strategy in sequence generation. 
Beam search is a \textbf{local search} that greedily optimizes the local objective at each step with constraints on search depth and beam width. % Greedy search is a special case 
It is used in many state-of-the-art NLP applications, including machine translation \cite{wu2016googles,ott-etal-2019-fairseq,wolf-etal-2020-transformers}, text summarization \cite{rush-etal-2015-neural,narayan-etal-2018-dont}, and image captioning \cite{anderson-etal-2017-guided}.
However, beam search is known to have high search error \cite{stahlberg-byrne-2019-nmt} due to the nature of local search. For example, \citet{welleck-etal-2020-consistency} reports that beam search can yield infinite-length outputs that the model assigns zero probability to. 

% \cite{Stahlberg2019} evaluated an exact search algorithm to find the most likely output according to the sequence model and showed that it performed poorly compared to beam search. The result suggests that search errors from beam search can mask model errors from the language model.

% UID hypothesis
% \citet{Meister2020} introduced the uniform information density (UID) hypothesis \cite{Levy2005, Levy2006} to explain this phenomenon. UID hypothesis claims that the people prefers to speak in a way that the average amount of surprisal for each token is evenly distributed among the utterances.
% They showed that information density is uniformly distributed by sequences generated by beam search.
% beam search is indeed optimizing the objective derived from the hypothesis.
% \citet{Meister2020} proposed multiple other regularizers to enforce the objectives derived from the UID hypothesis and showed that under these regularizers, beam search with large beam width does not significantly drop the performance, suggesting that such regularizer derived from the UID hypothesis is the source of the benefit of beam search.

% However, the constraint induced by beam search is not directly optimizing the objective derived by UID hypothesis. While UID hypothesis claims to distribute the information evenly, beam search constraint the surprisal to be greedily minimized.

%%%%%%%%%%%%%%%%%%%%%%%%%%%%%%%%%%%%%%
% Beam search and MAP decoding
Prior work has studied the two extreme ends of the search in terms of search depth. Beam search is a one-step local search without any consideration of the future step. Exhaustive search optimizes the global objective without regard to local optimality at each step.
Plenty of studies have investigated the effect of beam width on the search procedure and reported that a beam width that is neither too large nor too small is effective \cite{koehn-knowles-2017-six,stahlberg-byrne-2019-nmt,meister-etal-2020-beam}. However, in terms of search depth, little has been investigated between the two extreme ends. %, beam search and exhaustive search.
% only the two extreme ends, beam search and exhaustive search, have been investigated.
The research question we investigate is whether there is a better trade-off between the two ends in terms of search depth.

To analyze the effect of the search depth on the quality of the generated sequences, we introduce \textbf{\Analyzed{}} (\textbf{\ana{}}), a variant of beam search with multiple steps lookahead to improve the estimate of the next step. Beam search and exhaustive search is a special case of \ana{} with lookahead depth of $0$ and $\infty$, respectively.
We empirically evaluate the performance of \ana{} in machine translation tasks. 
The results show that \ana{} with up to 3-step lookaheads outperforms the performance of beam search and exhaustive search overall using Transformer-based models (Figure \ref{fig:locality}).
Although \ana{} outperforms beam search, the algorithm is computationally intensive. Therefore, we present \textbf{\Proposed{}} (\textbf{\pro{}}), a variant of beam search that heuristically simulates \ana{} with 1-step lookahead.
%Because optimization under the semi-local constraint is computationally intensive, we propose \Proposed{}, a simple search algorithm that approximate the search with 2-step semi-local constraint.
The algorithm is simple and requires only the same number of calls to the scoring function (probabilistic model) as beam search. %The advantages of our method is as follows:
It does not require any additional training or external modules. 
\pro{} is not restricted to specific tasks and is applicable to a wide range of natural language generation tasks, such as machine translation and text summarization.
We evaluate \pro{} for machine translation on WMT'14 En-Fr and En-De datasets \cite{Bojar2014} and for text summarization on CNN/DM \cite{hermann2015teaching} and XSum \cite{narayan-etal-2018-dont} datasets. %For translation, we evaluated on WMT'14 En-Fr and En-De datasets \cite{Bojar2014}. For summarization, we evaluated on CNN/DM \cite{hermann2015teaching} and XSum \cite{narayan-etal-2018-dont} dataset. 
The empirical results show that \pro{} improves upon vanilla beam search overall.

% \begin{itemize}
%     \item Simple: our algorithm is simple. It only needs a tweak on beam search.
%     \item Light-weighted: our algorithm requires no additional training or external modules. 
%     \item Domain-Independent: our algorithm is not restricted to specific task. As such, it is applicable to wide range of natural language generation tasks, such as translation and abstractive summarization.
% \end{itemize}

% For all the dataset, the proposed method outperformed vanilla beam search.
% In addition to the standard benchmarks, we evaluated our method on translation using a large language model. We used bloomz-mt \cite{}

% We also show that our approach is robust to beam width. While vanilla beam search fails with beam width gets large, the performance degradation with larger beam width is less significant for our method.

% \section{Related Work}

% % Beam Search
% Many people have investigated to improve the performance of the natural language generation.

% \todo cite thousands of papers on beam search

% \cite{Cohen} problem of beam search.
% \cite{Freitag2017} speeding up beam search.
% \cite{Hargreaves2021} incremental reranking method with trained model.
% \cite{Sridhar2022} reducing hallucination for summarization.

% UID hypothesis

\section{Neural Text Generation}

Sequence-to-sequence generation is the task of generating an output sequence $\vy$ given an input sequence $\vx$.
Probabilistic text generators define a probability distribution $p_\theta (\mathbf{y} | \mathbf{x})$ over an output space of hypotheses $\mathcal{Y}$ conditioned on an input $\mathbf{x}$.
The set of complete hypotheses $\mathcal{Y}$ is:
\begin{equation}
    \mathcal{Y} := \{\BOS \circ \mathbf{v} \circ \EOS | \mathbf{v} \in \mathcal{V}^*\},
\end{equation}
where $\circ$ is a string concatenation and $\mathcal{V}^*$ is the Kleene closure of a set of vocabulary $\mathcal{V}$. In practice, we set the maximum sequence length to $\nmax$ to limit the hypothesis space to $\vocab^{\nmax}$. The goal of decoding is to find the highest-scoring hypothesis for a given input. 

\subsection{Exhaustive Search}
One of the most important objectives is the maximum a posterior (MAP) objective to find the most probable hypothesis among all:
% \begin{equation}
%     \mathbf{y}^* := \argmax_{\mathbf{y} \in \mathcal{Y}} \texttt{score}(\vx, \vy).
% \label{eq:objective}
% \end{equation}
%
% The objective of the decoding is to find the most-probable hypothesis among all:
%
\begin{equation}
%    \mathbf{y}^{\text{MAP}} := \argmax_{\mathbf{y} \in \mathcal{Y}} \log p_\theta ( \mathbf{y} | \mathbf{x}).
    \mathbf{y}^* := \argmax_{\mathbf{y} \in \mathcal{Y}} \log p_\theta ( \mathbf{y} | \mathbf{x}).
\label{eq:map}
\end{equation}
We consider standard left-to-right autoregressive models for the model $p_\theta$:
\begin{equation}
    p_{\theta}(\mathbf{y} | \mathbf{x} ) = \prod_{t=1}^{|\mathbf{y}|} p_\theta (y_t | \mathbf{x}, \vy_{<t}).
\end{equation}
where each $p_\theta (y_t | \mathbf{x}, \vy_{<t})$ is a distribution with support over a set of vocabulary and the EOS: $\bar{\vocab} = \vocab \cup \{\EOS\}$.

% Decoding problem is a problem of solving the maximum a posteriori (MAP) objective (Eq. \ref{eq:map}) using a given autoregressive model $p_\theta$.
A straightforward solution to this problem is to maximize the MAP objective by exhaustively enumerating all possible hypotheses in $\mathcal{Y}$.
% This is known as maximum a posteriori (MAP) decoding.
Although it seems intuitive to use exhaustive search, prior work has pointed out several problems with this strategy. First, since the size of hypotheses set $|\mathcal{Y}|$ is extremely large, exhaustive search over $\mathcal{Y}$ is computationally infeasible. In fact, solving Eq. \ref{eq:map} is shown to be NP-hard \cite{chen-etal-2018-recurrent}. Second, even if we solve it optimally, the MAP objective often leads to low-quality results \cite{stahlberg-byrne-2019-nmt,Holtzman2020The,meister-etal-2020-beam}. % Therefore, decoding is performed almost exclusively with heuristic methods, such as beam search.

\subsection{Beam Search}

A common heuristic to solve the decoding problem is greedy search, a local search with a greedy procedure. 
Greedy search sequentially chooses the token $y_t$ at each time step $t$ that maximizes $p(y_t | \vy_{<t}, \vx)$ until the $\EOS$ token is generated or the maximum sequence length $\nmax$ is reached. Beam search is a generalization of greedy search where it selects the top $k$ tokens at each step.

Let $Y_t$ be the set of hypotheses at $t$-th step.
Beam search is expressed as the following recursion:
\begin{align}
    Y_0 &= \{\BOS\}, \nonumber\\
    Y_t &= \argtopk_{\vy \in \mathcal{B}_t} (\log p_\theta (\vy | \mathbf{x}) )
% Y_t &= \argmax_{Y' \subseteq \mathcal{B}_t, |Y'|=k} \log p_\theta (Y' | \mathbf{x}),
\label{eq:beam}
\end{align}
where the candidate set $\mathcal{B}_t$ is defined as:
\begin{equation}
    \mathcal{B}_t = \{\vy_{<t} \circ y_t | y_t \in \mathcal{\bar{V}} \land \vy_{<t} \in Y_{t-1} \},
\label{eq:localconstraint}
\end{equation}
for each $t > 0$. 
Beam search runs the recursion for a fixed number of iterations $\nmax$ and returns the set of hypotheses $Y_{\nmax}$. The most probable hypothesis (Eq. \ref{eq:map}) in $Y_{\nmax}$ is the output of the decoding.

Many of the decoding strategies used in statistical machine learning systems are variants of beam search \cite{vijayakumar2018diverse,meister-etal-2021-determinantal,anderson-etal-2017-guided,hokamp-liu-2017-lexically,king-etal-2022-dont,wan-etal-2023-faithfulness}. 
Although beam search does not solve Eq. \ref{eq:map} exactly, it is a surprisingly useful strategy for NLP models. In many settings, beam search outperforms exhaustive search in terms of downstream evaluation \cite{stahlberg-byrne-2019-nmt,Holtzman2020The,meister-etal-2020-beam}. 

The drawback of beam search is that it is known to have high search errors due to the nature of local search \cite{stahlberg-byrne-2019-nmt}.
For example, previous work has reported degenerations such as repetitions and infinite-length outputs \cite{Holtzman2020The,welleck-etal-2020-consistency}.

% Beam search is known to have high search error \cite{stahlberg-byrne-2019-nmt} due to the nature of local search.FFor example, W\citet{welleck-etal-2020-consistency} reports that beam search can yield infinite-length outputs that the model assigns zero probability to. 

\subsection{Uniform Information Density}
\citet{meister-etal-2020-beam} explains the effectiveness of beam search by introducing the Uniform Information Density (UID) hypothesis. The UID hypothesis claims that communicative efficiency is maximized when information is distributed as uniformly as possible throughout the sequence \cite{Levy2005,Levy2006}.
They study the information density of sentences generated by NMT systems quantitatively by measuring the amount of information conveyed by a word as surprisal \cite{hale-2001-probabilistic}.
The surprisal $u$ using a statistical language model is defined as follows:
\begin{align*}
    u_0(BOS) &= 0, \\
    u_t(y)   &= - \log p_\theta (y | \vx, \vy_{<t}).
\end{align*}
\citet{meister-etal-2020-beam} shows that the variance of surprisals and BLEU have a strong relationship in their empirical evaluation of NMT models.
They hypothesize that while restricting beam search leads to high search error in beam search, it also induces an inductive bias that may be related to promoting uniform information density, leading to the generation of higher quality sequences.

% \todo{cite a textbook for search algorithms to explain beam search as a heuristic search method} \\
% Beam search is a local search.

% Beam search optimizes the MAP objective with a local constraint which restrict each token of the sequence to be top-$k$ (Eq. \ref{eq:beam}).
% Beam search optimizes the MAP objective with a local search which 

%Beam search is central to many of natural language generation tasks. Many of the decoding strategies used in statistical machine translation (SMT) are variants of beam search. Beam search typically has large search errors, yet it typically generates coherent text from probabilistic models.

% \section{\Analyzed{}}
\section{Analysis of Search Depth}
To study the effect of search depth on the properties of the decoding strategy, we first introduce \Analyzed{} (\ana{}), a generalization of beam search and exhaustive search that can control the search depth (Section \ref{sec:analyzed}).
Then, we empirically evaluate the effect of the search depth on the quality of the output sequence using \ana{} (Section \ref{sec:analysis}).

\subsection{\Analyzed{}}
\label{sec:analyzed}

\begin{figure}
    \centering
    \includegraphics[width=\columnwidth]{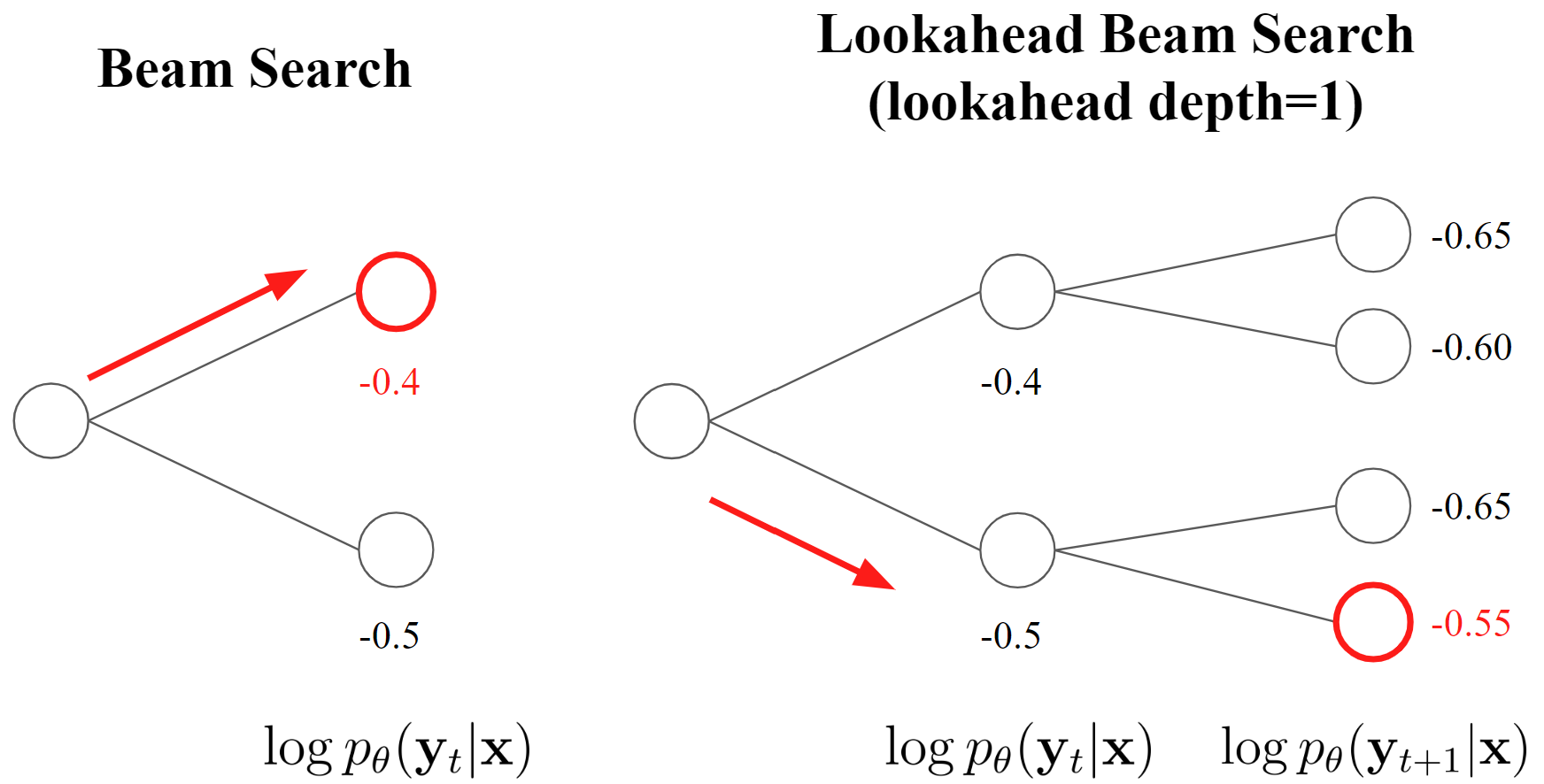}
    \caption{Comparison of \Analyzed{} and beam search. While beam search chooses the next hypotheses according to the current score of the hypothesis, \analyzed{} chooses them according to the current score plus the highest possible score achievable within $d$-step future.}
    \label{fig:analyzed}
\end{figure}

% \todo{What's the best name for the algorithms? it's rather a semi-greedy and local search, not semi-local search. but "greedy" is a characteristic of the procedure whereas "local" is the characteristic of the optimization problem. so "local" is more to the problem. but locality comes from the greedy procedure, so...}
We propose \Analyzed{} (\ana{}), a simple extension of beam search.
\ana{} deploys a lookahead strategy to optimize the multi-step score instead of the immediate score (Figure \ref{fig:analyzed}).
In addition to the score given by the current partial hypothesis, \ana{}-$d$ incorporates the maximum possible score achievable in the $d$-step future. We replace Eq. \ref{eq:beam} with the following:
\begin{align}
    Y_0 &= \{\BOS\}, \nonumber\\
    Y_t &= \argtopk_{\vy \in \mathcal{B}_t} (\log p_\theta (\vy | \mathbf{x}) + h_d(\vy) ),
    % Y_t &= \argtopk_{\vy \in \mathcal{B}_t} (h_m(\vy)), \nonumber\\
\label{eq:lookahead}
\end{align}
% where $\vy_{1:t+m}$ is a hypothesis concatenating $m$ tokens to $\vy$: $\vy_{1:t+m} = \vy \circ y_{t+1} \circ ... \circ y_{t+m}$.
where $h_d(\vy)$ is the highest score achievable of $d$-step future starting from $\vy$. $h_d(\vy)$ is defined as:
\begin{align}
    h_d(\vy_{1:t}) &= \max_{\vy_{1:t+d} \in \mathcal{B}_t^d} \log p_\theta (\vy_{1:t+d} | \vx, \vy), \nonumber\\
    \mathcal{B}_t^d &= \{\vy_{1:t} \circ y_{t+1} \circ ... \circ y_{t+d} | y_{t+1}, ..., y_{t+d} \in \bar{\vocab}\}.
\label{eq:lookahead-estimate}
\end{align}
The lookahead depth $d$ is the hyperparameter of the algorithm to control the locality of the search. 
The search becomes more local and shallow as $d$ becomes smaller. In particular, if $d=0$, it recovers beam search.
The search becomes more exhaustive with larger $d$, and $d \geq \nmax$ recovers exhaustive search.
% \begin{equation}
%     h_m(\vy) = \max_{\vy'_{t+1:t+m} \in \bar{\vocab}^m} \log p_\theta (\vy'_{t+1:t+m} | \vy, \vx).
% \label{eq:lookahead-estimate}
% \end{equation}

% $\vy_{1:t+d}$ is a hypothesis concatenating $d$ tokens to $\vy$: $\vy_{1:t+d} = \vy \circ y_{t+1} \circ ... \circ y_{t+d}$.
% $h_d$ is the return of the $d$-lookahead policy with respect to the scoring function \cite{NEURIPS2020_a18aa23e}. \todo{I'm not willing to stress the novelty of the method...}

% While beam search selects the next token only considering the score of the immediate next step, \ana{} selects the top-$k$ options with highest $d$-lookahead returns (Figure \ref{fig:analyzed}).
% \todo{\ana{} is a beam search version of $h$-RTDP, an online planning algorithm with lookahead policy where it computes top-$k$ candidates instead of the top candidate \cite{NEURIPS2020_a18aa23e}.}

\begin{proposition}
    \Analyzed{} (\ana{}) is a generalization of beam search and exhaustive search. That is,
    \begin{enumerate}
        \item \ana{}-$0$ recovers beam search.
        \item \ana{}-$d$ with $d \geq \nmax$ recovers exhaustive search. 
    \end{enumerate}
\end{proposition}
The proof is immediate from the definition of \ana{}.

% Although beam search takes local constraint, it only takes into account of immediate neighbor tokens. 
% Under current formulation, local constraint only considers the constraints between the immediate neighboring tokens.

% While the local constraint of beam search is one of the ways to enforce the uniformity of the information density, it is not directly optimizing UID. In fact, it is too restrictive to force the constraint to every token.

% To investigate the effect of the local constraint, we introduce a semi-local constraint that generalize the local constraint of the beam search.
% $m$-step semi-local constraint enforces that the top-$k$ tokens are selected at least every $m$-steps. Search under $m$-step semi-local constraint requires that for every step $t$, the candidate set is as follows:

% \begin{equation}
%     \mathcal{B}^m_t = \mathcal{B}^{m-1}_t \cup \{\mathbf{y}_{t-m} \circ \vy_{m} | \vy_{m} \in \mathcal{\bar{V}}^m \land \vy_{t-m} \in Y_{t-m} \},
% \label{eq:semilocalconstraint}
% \end{equation}

% The semi-local constraint is a generalization of the beam search and MAP decoding.

% \begin{lemma}
%     Semi-local search with $m=1$ recovers beam search. Semi-local search with $m \geq \nmax$ recovers the map decoding.
% \end{lemma}

A straightforward implementation to compute $h_d(\vy)$ is breadth-first search. However, it needs to call the scoring function for $k | \bar{\vocab}|^d$ times per each step. This is prohibitively expensive because the vocabulary size $|\vocab|$ is large in many tasks (e.g. $>$30000).
To reduce the computation time, we implement the evaluation of $h_d$ by best-first branch-and-bound search. See Appendix for details. Note that the algorithm is guaranteed to find the same $h_d$ as the breadth-first search, thus preserving the result of the decoding. 

% List \ref{lst:analyzed} describes the procedure of the best-first branch-and-bound. \todo{Put it in Appendix?} \todo{It can be as simple as dijkstra, why bother explaining it?}

% \begin{listing}
% \caption{Best-First BnB for \Analyzed{}}
% \label{lst:analyzed}
% \renewcommand{\algorithmicrequire}{\textbf{Input:}}
% \renewcommand{\algorithmicensure}{\textbf{Output:}}
% \begin{algorithmic}[1]
% \REQUIRE a set of hypotheses $Y_{t-1}$ of length $t-1$ \\
% \ENSURE a set of hypotheses $Y_t$ of length $t$
% \end{algorithmic}
% \end{listing}

% \section{Analysis of Search Depth}
\subsection{Empirical Evaluation}
% \section{Analysis of Search Locality}
\label{sec:analysis}

%%%%%%%%%%%%%%
% Purpose 
% We explore how the lookahead strategy in text generated by neural probabilistic language models affects its quality. To this end, 
To empirically investigate the effect of search depth on the output sequence, we decode neural machine translation (NMT) models using \ana{}. 
%%%%%%%%%%%%%%
% What to measure
We evaluate the text quality by BLEU \cite{papineni-etal-2002-bleu} using the SacreBLEU system \cite{post-2018-call}. 
% \todo do we mention UID hypothesis? if not, remove this.
%%%%%%%%%%%%%%
% Dataset and Code
Experiments are performed on WMT'14 En-Fr and WMT'14 En-De datasets \cite{Bojar2014}. For reproducibility, we use the pretrained models provided by fairseq \cite{ott-etal-2019-fairseq}.\footnote{\url{https://github.com/facebookresearch/fairseq/tree/main/examples/translation}}
We build the decoding framework in SGNMT \cite{stahlberg-etal-2017-sgnmt}.\footnote{\url{https://github.com/ucam-smt/sgnmt}} 
Due to the long duration (Table \ref{tab:wtime}) and computational constraints, we present the evaluation on the first 100 sentences.
% We report the evaluation results on the first 100 sentences due to the extensive duration (Table \ref{tab:wtime}) and computational constraints.
The results on WMT'14 En-De and the other ablation studies are reported in Appendix.

\begin{figure}
    \centering
    \includegraphics[width=\columnwidth]{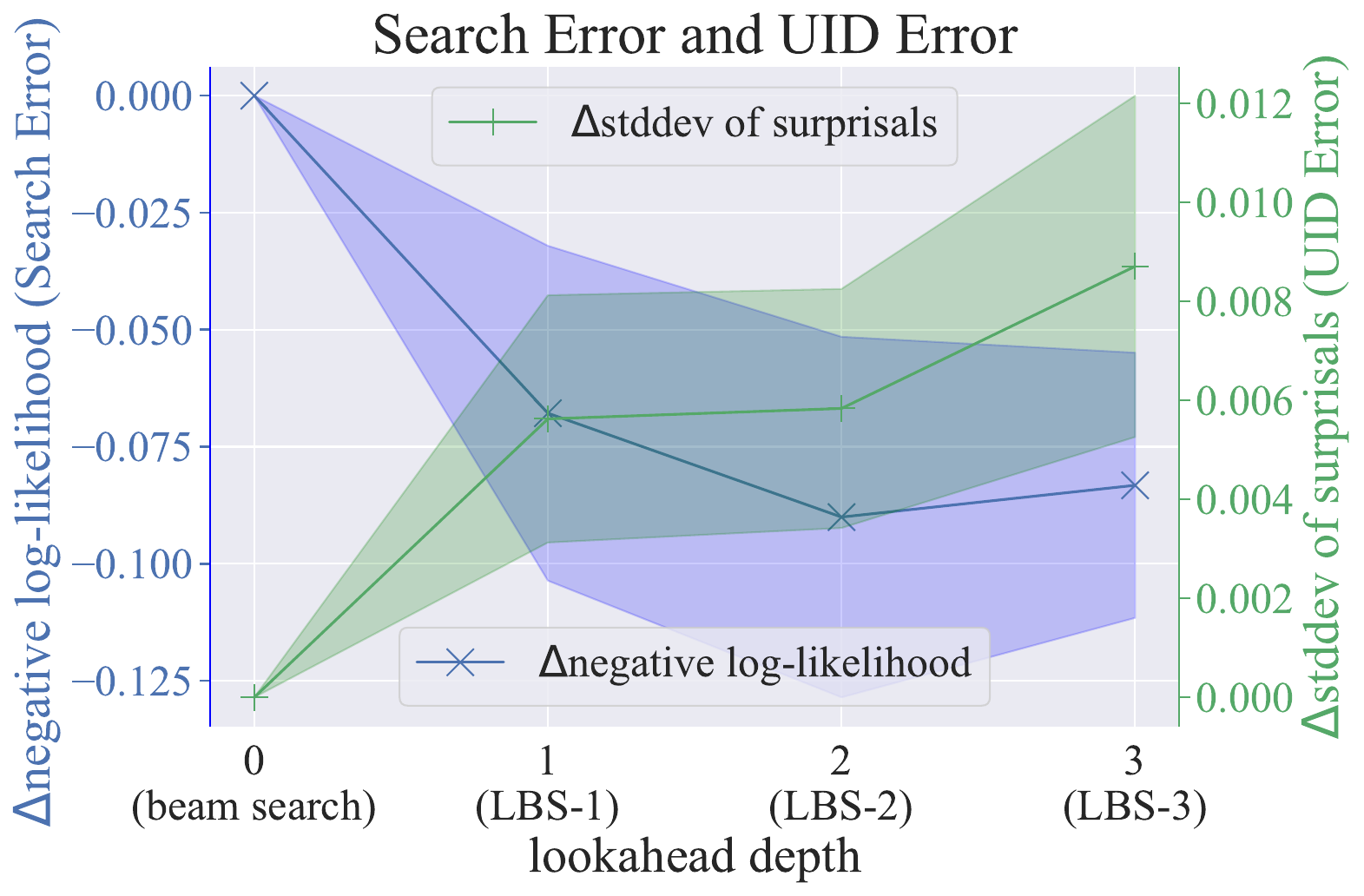}
    \caption{Difference in negative log-likelihood (search error) and average standard deviation of surprisals per sequence (UID error) of \analyzed{} (\ana{}) compared to beam search. The bold line represents the mean over the beam widths. The shaded area shows the standard error.}
    \label{fig:search-uid-depth}
\end{figure}

\begin{figure}[htb]
    \centering
    \includegraphics[width=\columnwidth]{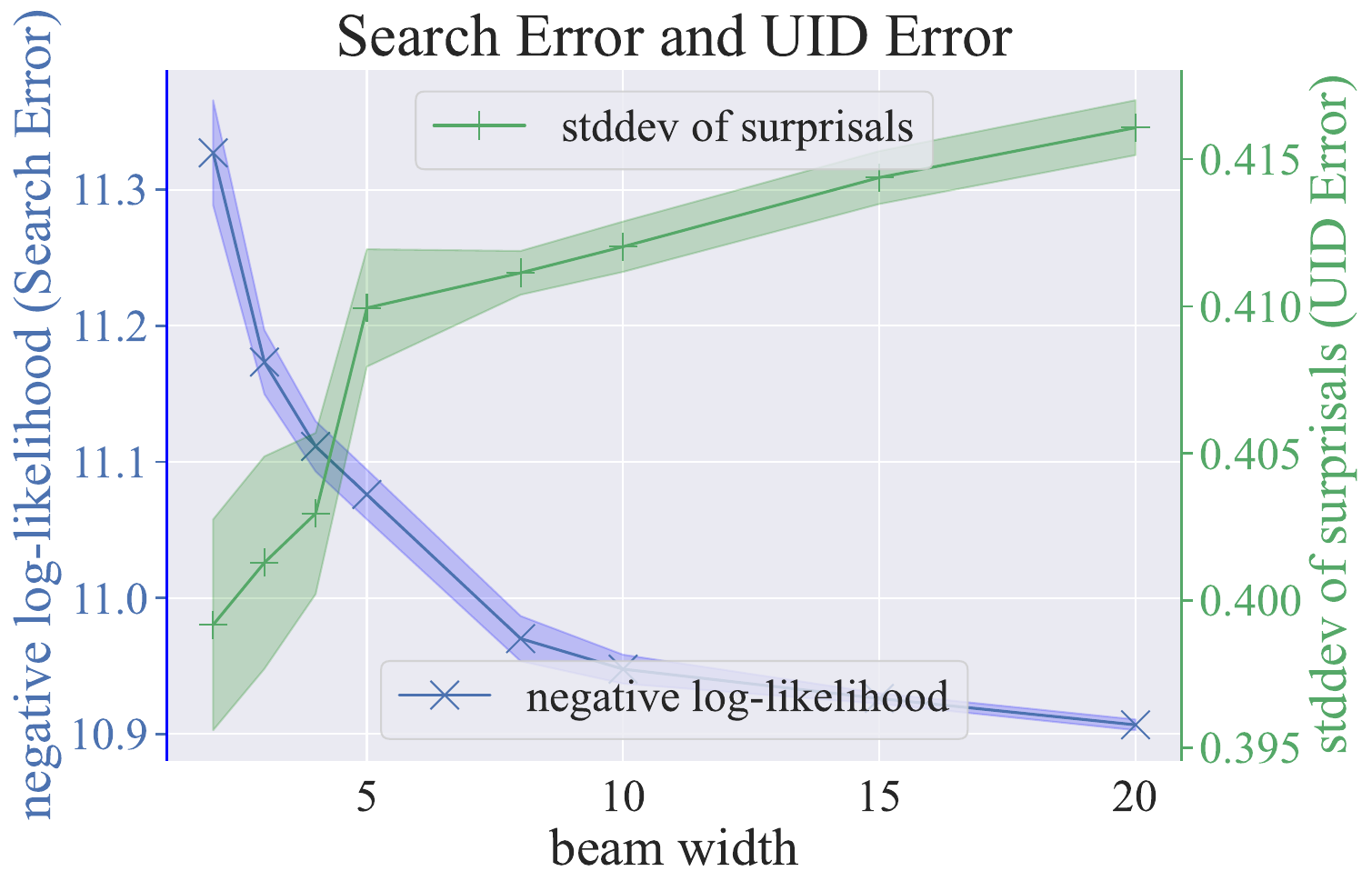}
    \caption{Negative log-likelihood (search error) and the average standard deviation of surprisals per sequence (UID error) by \analyzed{} (\ana{}). The bold line represents the mean over lookahead depth of $d=0, 1, 2,$ and $3$. The shaded area shows the standard error.}
    \label{fig:search-uid-width}
\end{figure}

The summary of the analysis is as follows.
\begin{itemize}
    \item \ana{} achieves higher BLEU scores compared to beam search (Figure \ref{fig:locality}).
    \item We observe a trade-off between \textbf{search error} and \textbf{UID error} for both beam width and lookahead depth (Figure \ref{fig:search-uid-depth} and \ref{fig:search-uid-width}).
    \item Lookahead depths of up to 3 have little effect on sequence length, while beam width has a strong negative correlation with it (Figure \ref{fig:length}).
\end{itemize}

% \todo{we also have the result of last 50 sentences. does it matter?}

%\todo need to explain best-first branch-and-bound for NLP researchers? Put it to Appendix?: 

\subsubsection{BLEU Score}
Figure \ref{fig:locality} demonstrates how the lookahead strategy affects the quality of the results as the lookahead depth varies. In particular, \ana{}-2 achieves the best overall BLEU score. We observe a reduced improvement with a lookahead depth of 3 (\ana{}-3) compared to \ana{}-2. As reported in previous work \cite{stahlberg-byrne-2019-nmt}, exhaustive search (i.e. \ana{}-$\infty$) performs very poorly (See Appendix).

% \begin{figure}[htbp]
%   \begin{minipage}[b]{0.48\linewidth}
%     \centering
%     \includegraphics[width=\columnwidth]{depth-perplexity.pdf}
%     \caption{Average perplexity per sentence according to the model that is being generated from.}
%     \label{fig:perplexity}
%   \end{minipage}
%   \begin{minipage}[b]{0.48\linewidth}
%     \centering
%     \includegraphics[width=\columnwidth]{depth-ustddev.pdf}
%     \caption{Average standard deviation of surprisals per sequence (UID error) and BLEU with \analyzed{} (\ana{}) for different beam widths and lookahead depths.}
%     \label{fig:uid}
%   \end{minipage}
% \end{figure}

\subsubsection{Why is there a ``sweet spot'' for lookahead depth?}
We observe that a lookahead depth of $d=2$ outperforms $d=0, 1,$ and $3$ (Figure \ref{fig:locality}). The question is why there is a ``sweet spot'' for lookahead depth. 
Our hypothesis is that this phenomenon can be explained by the trade-off between the \textit{search error} and the \textit{UID error}. 
% We hypothesize that this can be explained by the trade-off of \textit{search error} and \textit{UID error}. %, as is hypothesized for beam width \cite{meister-etal-2020-beam}.
The search error is measured as the loss of negative log-likelihood compared to beam search (Figure \ref{fig:search-uid-depth}, left axis). 
We observe that increasing the lookahead depth reduces the search error, as measured by the negative log-likelihood. 
A prior study reports that the deviation from uniform information density measured by the standard deviation of surprisals has a negative correlation with the BLEU score \cite{meister-etal-2020-beam}. We report the standard deviation of surprisals as UID error in Figure \ref{fig:search-uid-depth} (right axis). We observe a negative correlation between lookahead depth and the standard deviation of surprisals.

Overall, we observe that deeper lookaheads improve the search error, but at the cost of higher UID error at the same time. 
These results suggest that a lookahead depth of 2 happens to be a better trade-off between search error and UID error in our experimental setting. % Similarly, the tradeoff between the two errors can be an explanation of the reason why beam widths of 3-10 are known to be more effective than smaller or larger beam widths, though it may be because of the reduced average sequence length (Figure \ref{fig:length}).
We also observe a similar trend for beam width (Figure \ref{fig:search-uid-width}), as indicated by \citet{meister-etal-2020-beam}. % As beam width increases, search error decreases but UID error increases. 

\begin{figure}[tb]
    \centering
    \includegraphics[width=\columnwidth]{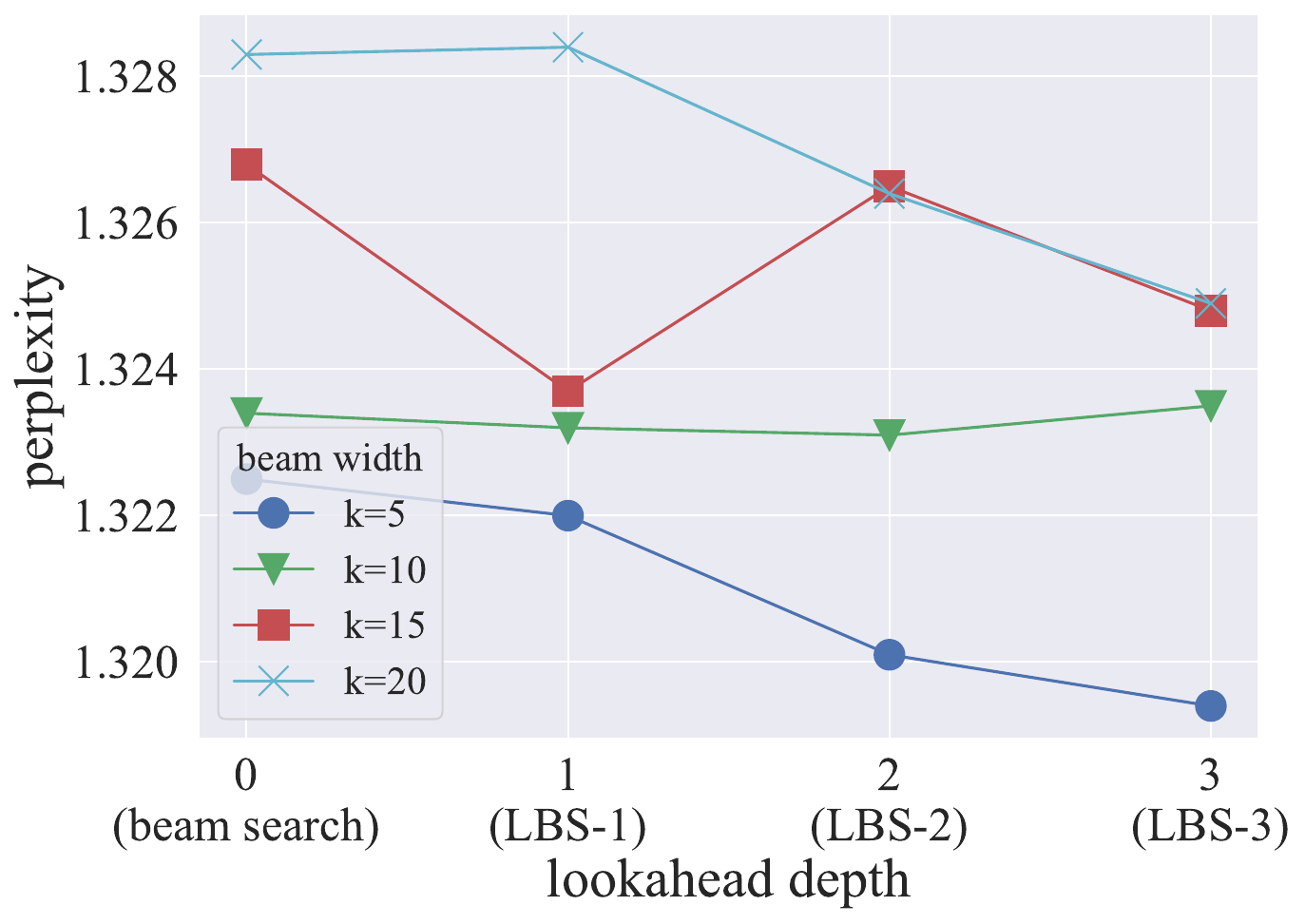}
    \caption{Average perplexity per sentence according to the model that is being generated from.}
    \label{fig:perplexity}
\end{figure}

Figure \ref{fig:perplexity} reports the average perplexity. We observe that increasing the lookahead depth tends to improve the perplexity. Therefore, search error, measured as both negative log-likelihood and perplexity, decreases with increasing lookahead depth. Thus, search error \textit{alone} does not explain why $d=2$ has the highest BLEU score.

\begin{figure}[tb]
    \centering
    \includegraphics[width=\columnwidth]{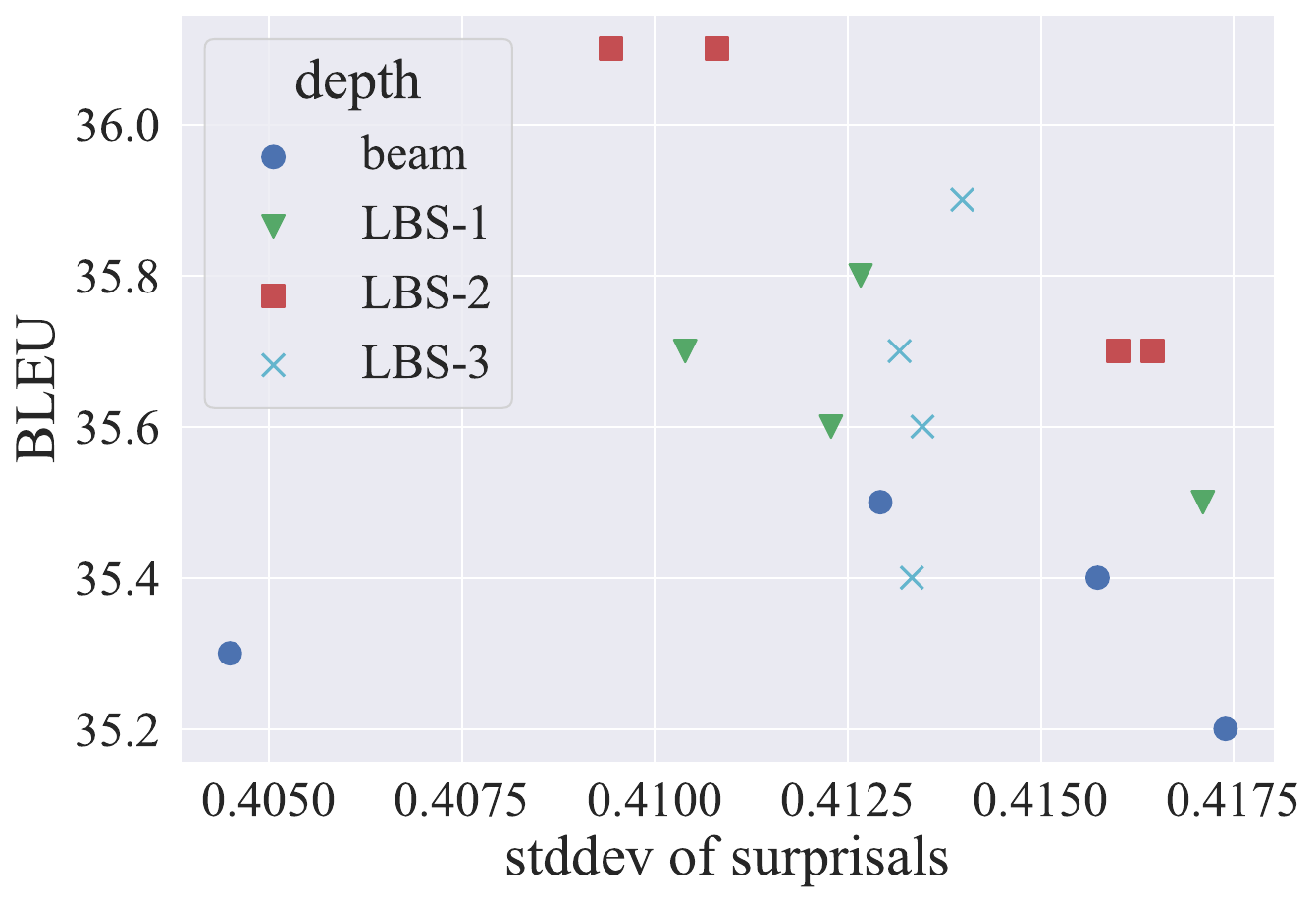}
    \caption{Average standard deviation of surprisals per sequence (UID error) and BLEU with \analyzed{} (\ana{}) for different beam widths and lookahead depths.}
    \label{fig:uid}
\end{figure}

% \subsubsection{UID Error}
Figure \ref{fig:uid} shows the standard deviation of surprisals and BLEU for different numbers of lookahead depths and beam widths. Although \ana{} has higher BLEU scores than beam search, it also has a higher average standard deviation of surprisals per sentence. 
Therefore, the UID error \textit{alone} cannot account for the effect of lookahead depth on the BLEU scores.

\begin{figure}[tb]
    \centering
    \includegraphics[width=\columnwidth]{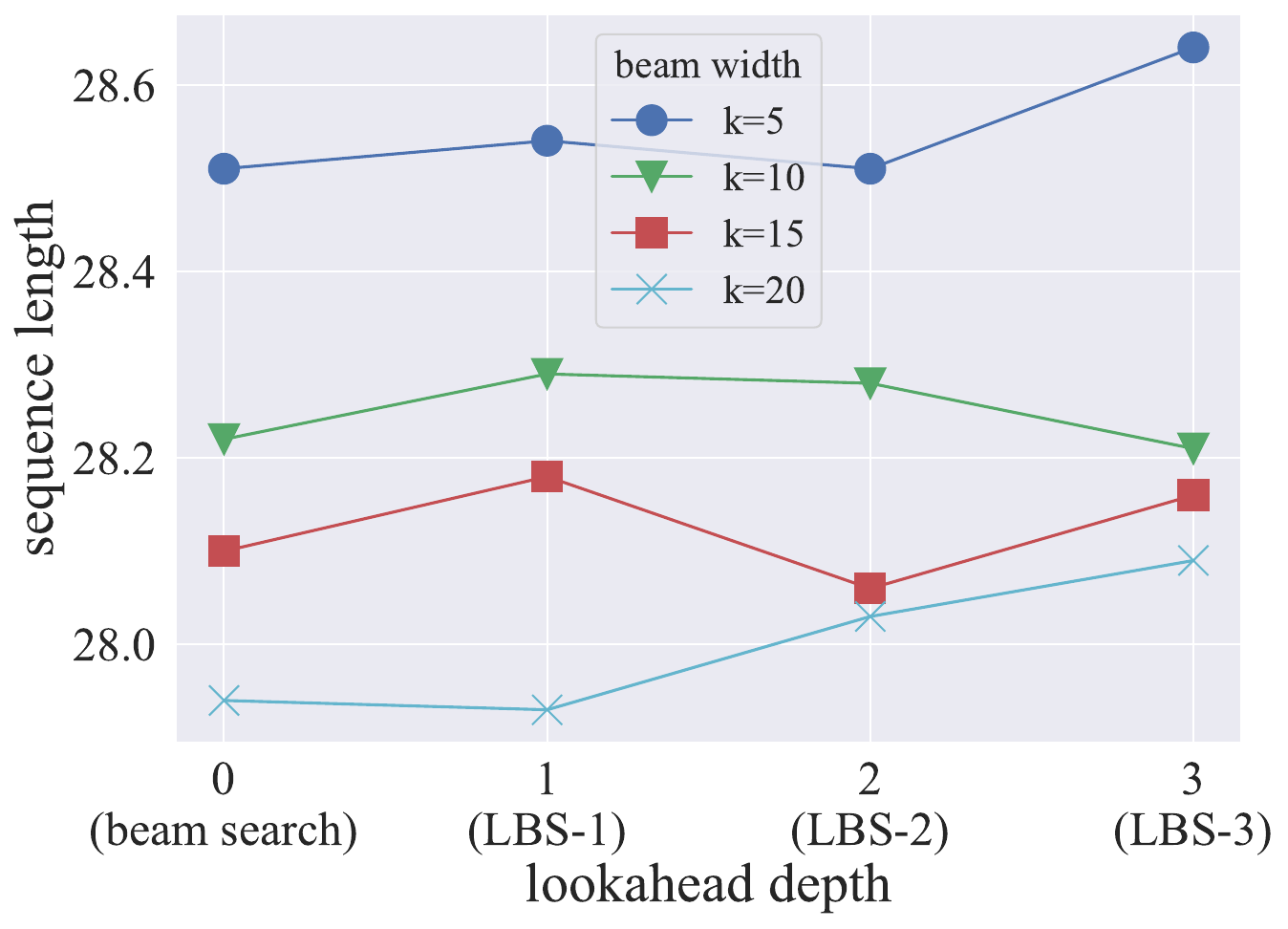}
    \caption{Average sequence length for varying lookahead depth and beam width. The correlation of beam width and lookahead depth with the average sequence length is $-0.92$ and $0.12$, respectively.}
    \label{fig:length}
\end{figure}

\subsubsection{Sequence Length}
Figure \ref{fig:length} shows the average length of the output sequence. 
We observe that while widening the beam reduces the output sequence length, deepening the lookahead by up to 3 steps does not. The correlation of beam width and lookahead depth with sequence length is $-0.92$ and $0.12$, respectively. While beam width has a clear negative correlation with output sequence length, lookahead depth has little effect on sequence length. Thus, average sequence length alone does not explain the effect of lookahead depth on BLEU score.
% \todo{report length of MAP decoding by our experiments.} While \citet{stahlberg-byrne-2019-nmt} reports $>$50 percent of the sequence being non-existent using MAP decoding, we observe none of the sequence being non-existing using \ana{} in our experiments.

\subsubsection{Running Time}
We report the wall-clock time of \ana{} in Table \ref{tab:wtime}. \ana{} is significantly slower than beam search especially when the lookahead depth is large. The wall time is roughly proportional to the number of calls to the scoring function (see Appendix). 
Note that the wall-clock time is hardware dependent. All the experiments are performed on \texttt{g4dn.xlarge} instances on AWS EC2 (4 vCPU cores, 16 GB memory, and an NVIDIA T4 GPU).
% \todo{wall-clock time vs. number of calls to the score function? walltime is more immediately important to practitioners.}

\begin{table}%[htb]
    \centering
    \begin{tabular}{c|rrrr}
    \toprule
    Decoder & $k=5$ & $k=10$ & $k=15$ & $k=20$ \\
    \hline \hline
    %%%%%%%%%%
    % Results on 100 sentences
    %%%%%%%%%%    
    beam & 2.04 & 3.98 & 5.44 & 7.59 \\
    \ana{}-1 & 22.93 & 58.55 & 90.45 & 138.31 \\
    \ana{}-2 & 53.91 & 165.13 & 302.62 & 482.58 \\
    \ana{}-3 & 102.71 & 329.69 & 640.80 & 999.73 \\
    %%%%%%%%%%
    % Results on 400 sentences
    %%%%%%%%%%
    % beam & 1.86 & 3.90 & 5.94 & 7.91 \\
    % \ana{}-1 & 22.59 & 55.83 & 98.23 & 144.26 \\
    % \ana{}-2 & 53.20 & 169.63 & 310.93 & 499.51 \\
    % \ana{}-3 & 104.08 & 338.60 & 649.10 & 1057.85 \\
    \bottomrule
    \end{tabular}
    \caption{Average running time (sec) per sentence.}
    \label{tab:wtime}
\end{table}

\section{\Proposed{}}

\begin{listing}
\caption{\Proposed{} (\pro{})}
\label{lst:proposed}
\renewcommand{\algorithmicrequire}{\textbf{Input:}}
\renewcommand{\algorithmicensure}{\textbf{Output:}}
\begin{algorithmic}[1]
\REQUIRE a set of hypotheses $Y_{t-1}$ of length $t-1$ \\
\ENSURE a set of hypotheses $Y_t$ of length $t$
\STATE $\mathcal{B}'_{t, 0} = \emptyset$ \\
\STATE $\{\vy_{t-1}^1, \vy_{t-1}^2,...,\vy_{t-1}^k\} = \text{sort}(Y_{t-1})$ in a descending order of $p(\vy_{t-1}^i | \vx)$
\FOR {$i \in \{1..k\}$}
    % \STATE $\mathcal{B}(\vy_{t-1}^i) = \{\vy_{t-1}^i \circ y | y \in \mathcal{\bar{V}}\}$ \\
    % \STATE $\mathcal{B}_{t, i} = \mathcal{B}'_{t, i-1} \cup Y_{t}^i$ \\
    \STATE $\mathcal{B}_{t, i} = \mathcal{B}'_{t, i-1} \cup \{\vy_{t-1}^i \circ y | y \in \mathcal{\bar{V}}\}$ \\
    \STATE $\vy_{t}^i = \argmax_{\vy \in \mathcal{B}_{t, i}} \log p_\theta (\vy | \vx)$
    \STATE $\mathcal{B}'_{t, i} = \mathcal{B}_{t, i} \setminus \{\vy_{t}^i\}$
\ENDFOR
\RETURN $Y_t = \{\vy_{t}^i, \vy_{t}^2,...,\vy_{t}^k\}$
\end{algorithmic}
\end{listing}

% In Section \ref{sec:analysis} we observe that an optimal lookahead depth may be larger than 0.

While \ana{} outperforms beam search, it requires too much computation for sequence generation tasks with large vocabulary sizes (Table \ref{tab:wtime}).
We present \Proposed{} (\pro{}), a variant of beam search that heuristically simulates \ana{}-1 that runs in the same amount of computation as beam search $O(\nmax k |\vocab| )$. 
\pro{} simulates \ana{} by {\it looking behind} to consider the past score $p(\vy_{<t} | x)$ in addition to the current score. 

% \begin{align}
%     Y_t &= \argtopk_{\vy \in \mathcal{B}_t} (\log p_\theta (\vy | \vx) + h_1(\vy)) \nonumber \\
%     &= \argtopk_{\vy \in \mathcal{B}_t} (\log p_\theta (\vy | \vx) + \log p_\theta(\vy_{1:t+1} | \vx, \vy )) \\
%     &= \argtopk_{\vy \in \mathcal{B}_{t'-1}} (\log p_\theta (\vy_{1:t-1} | \vx) + \log p_\theta(\vy_{1:t} | \vx, \vy_{1:t-1})) \\
% \end{align}

% By $t' = t+1$

% \begin{align}
%     \vy_t &= \argtopi_{\vy \in \{\vy_{1:t-1} \circ y | \vy_{1:t-1} \in Y_{t-1}^i\}} (\log p_\theta (\vy_{1:t-1} | \vx) + \log p_\theta(y_t | \vx, \vy_{1:t-1}) ), \\ 
%     Y_{t-1}^i &= \argtopi_{\vy_{1:t-1} \in Y_{t-1}} (\log p_\theta (\vy_{1:t-1} | \vx))
% \end{align}

\begin{figure}[tb]
    \centering
    \includegraphics[width=\columnwidth]{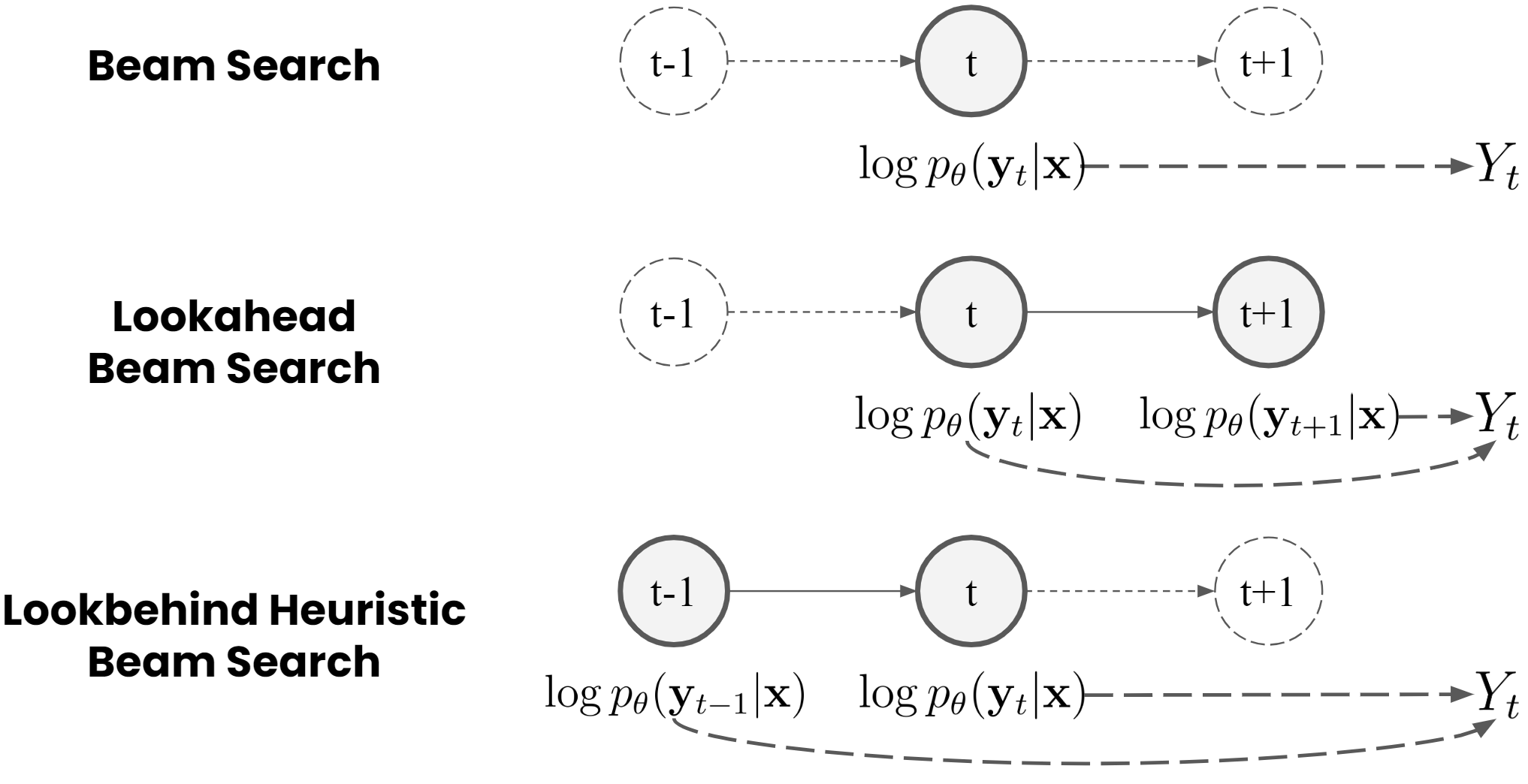}
    \caption{Conceptualized visualization of \Proposed{}. \ana{} uses the current score and the future score to select the next set of hypotheses $Y_t$ whereas \pro{} uses the current score and the previous score.}
    \label{fig:method}
\end{figure}

While beam search determines the next set of hypotheses $Y_t$ considering the score of the current sequence $\log p_\theta (\vy_t | \vx)$, \ana{} with 1-step lookahead (\ana{}-1) selects $Y_t$ according to both the current score ($\log p_\theta (\vy_t | \vx)$) and the 1-step future score ($h_1(\vy) = \max \log p_\theta (y_{t+1} | \vx, \vy)$).
Since computing $h_1(\vy)$ requires a computation of the probabilities of succeeding-step, \ana{}-1 is computationally expensive, with a complexity of $O(\nmax k |\vocab|^2)$ in total.
\pro{} applies two heuristics to \ana{}-1 to reduce the computational overhead. First, we use $\log p_\theta (\vy_{t-1} | \vx)$ to account for the two successive scores (Figure \ref{fig:method}). Since $\log p_\theta (\vy_{t-1} | \vx)$ is already computed in the beam search procedure, there is no additional overhead. Second, we limit the number of hypotheses to be considered to $k$, while heuristically retaining the most promising ones (Proposition 2). This reduces the computational complexity of the algorithm to $O(k |\vocab|)$ for each step.

% The procedure of \pro{} is depicted in List \ref{lst:proposed}. 
% We order the hypotheses in descending order of the score $p(\vy_i | x)$ (Line 2) and consider each specific hypothesis in the beam sequentially. 
% We give priority to the hypothesis with the higher score at the $t-1$ step by the following procedure.
\pro{} gives priority to the hypothesis with the higher score at step $t-1$ by ordering the hypotheses by their scores (Line 2 in Algorithm \ref{lst:proposed}). It considers each specific hypothesis in the beam sequentially (Line 3).
We maintain a priority queue $\mathcal{B}_{t, i}$ containing only children of hypotheses at the current beam slot and slots before it (Line 4). The highest scoring candidate is then taken from $\mathcal{B}_{t, i}$ and stored as $\vy_{t}^i$ (Lines 5 and 6).\footnote{In many implementations, including fairseq and HuggingFace's Transformers library, beam search takes top-$2k$ hypotheses instead of top-$k$. This allows it to pick the first $k$ of those that don't predict $\EOS$ to continue with. We follow this implementation and pop the top-$2$ hypotheses in Lines 5 and 6 to generate $2k$ hypotheses.} 

Formally, we denote $\vy_{t}^i$ as the $i$-th hypothesis of the hypotheses set $Y_t$ where $i \in \{1..k\}$.

\begin{align}
    Y_{0} &= \{\BOS\}, \\
    Y_t &= \textsc{Sort}(\{\vy_{t}^1,...,\vy_{t}^k\}), \\
\end{align}
where $\vy_{t}^i$ is generated as follows:

\begin{equation}
    \vy_{t}^i = \argmax_{\vy' \in B_{t, i}} \log p_\theta (\mathbf{y'} | \mathbf{x}).
\end{equation}
The candidate set $\mathcal{B}_{t, i}$ is defined as:

\begin{equation}    
    \mathcal{B}_{t, i} = \{\mathbf{y}_{t-1} \circ y | y \in \bar{\vocab} \land  \mathbf{y}_{t-1} \in Y_{t-1}^{1:i}\}  \setminus \{Y_{t}^{1:i-1}\},
\label{eq:candset}
\end{equation}
where $Y_{t}^{1:i} = \{\vy_{t}^1,...,\vy_{t}^i\}$.
By prioritizing hypotheses deemed promising at the $t-1$ step, \pro{} retains some of these hypotheses in the subsequent $t$-th step, even if they are not top-$k$ at the $t$-th step. This is in contrast to beam search, which eliminates all non-top-$k$ hypotheses at each iteration.
% Formally, at $t$-th iteration of the \Proposed{}, $i$-th candidate hypothesis $\vy_{t}^i = \vy_{t-1}^i \circ y_t$ is guaranteed that (1) $\vy_{t-1}^i$ is the top-$i$ candidate amongst $Y_{t-1}$ and (2) $\vy_{t}^i$ is the top-$i$ candidate amongst $\mathcal{B}_{t}^i$.
More formally, this procedure guarantees the following properties:
\begin{proposition}
    At $t$-th iteration of the \Proposed{}, $i$-th candidate hypothesis $\vy_{t}^i = \vy_{t-1}^i \circ y_t$ is guaranteed that 
    \begin{enumerate}
        \item $\vy_{t-1}^i$ has the top-$i$ highest score among $Y_{t-1}$
        \item $\vy_{t}^i$ has the top-$i$ highest score among $\mathcal{B}_{t}^i$.
    \end{enumerate}
\end{proposition}
The proof is immediate from the construction (Algorithm \ref{lst:proposed}).

% \todo{how is LHBS related to LBS?}
The cost of retaining non-optimal hypotheses at the $t$-th step is that we have fewer top-$k$ hypotheses at the $t$-th step. In other words, we seek to deepen the search at the cost of search width $k$. 
We hypothesize that it is an effective strategy to take depth over width from our analysis (Section \ref{sec:analysis}). % and from numerous studies showing that having a beam width that is too large can adversely affect the quality of the results \cite{koehn-knowles-2017-six,murray-chiang-2018-correcting,yang-etal-2018-breaking,stahlberg-byrne-2019-nmt,pmlr-v97-cohen19a}.

% \section{Evaluation of \Proposed{}}
\section{Experiments}

We evaluate \proposed{} (\pro{}) in machine translation and text summarization. For all experiments we use publicly available pretrained models and datasets.

\subsection{Machine Translation}

% We implemented \Proposed{} to two neural machine translation systems: fairseq \cite{ott-etal-2019-fairseq} and Joey NMT \cite{kreutzer-etal-2019-joey}.
To evaluate the performance of \pro{}, we decode NMT models using the proposed decoding strategy.
The experiments are performed on the full WMT'14 En-Fr and WMT'14 En-De test datasets \cite{Bojar2014}. 
For reproducibility, we use the pretrained models made publicly available by fairseq library \cite{ott-etal-2019-fairseq}.
%We used the same hyperparameter as in prior literature.
We use BLEU \cite{papineni-etal-2002-bleu} to evaluate the text quality using the fairseq-score tool. 
All models and data information can be found in the fairseq repository.\footnote{\url{https://github.com/facebookresearch/fairseq/tree/main/examples/translation}}
On both datasets and across different beam widths (except for En-Fr with $k=5$), the results indicate that the proposed method produces higher quality sequences (Table \ref{tab:translate}).

\begin{table}
    \centering
\begin{tabular}{c|c|ccc}\toprule
Dataset & Decoder & $k=5$ & $k=10$ & $k=20$ \\ \hline\hline
\multirow{2}{*}{En-Fr} & beam & \textbf{42.98} &42.95 &41.57 \\ 
 & \pro{} & 42.97 & \underline{\textbf{43.01}} &\textbf{41.64} \\ \hline
\multirow{2}{*}{En-De} & beam & 29.17 &29.14 &28.95 \\
 & \pro{} & \underline{\textbf{29.29}} &\textbf{29.25} &\textbf{28.97} \\
% WMT'20 Ja-En & beam & 17.43 &17.51 &20.06 \\ \hline
% WMT'20 Ja-En & \pro{}  & \textbf{17.74} &\textbf{17.75 }&\textbf{20.25} \\
\bottomrule
    \end{tabular}
    \caption{BLEU scores for predictions generated with \Proposed{} (\pro{}) and beam search (WMT'14 En-Fr). The best scores per beam size are bolded. The best score is underlined.}
    \label{tab:translate}
\end{table}

% \subsubsection{Evaluation on Joey NMT}
% \begin{figure}
%     \centering
%     \includegraphics[width=\columnwidth]{wmt14ende-joeynmt.png}
%     \caption{wmt14 ende joeynmt}
%     \label{fig:ende}
% \end{figure}

% \begin{figure}
%     \centering
%     \includegraphics[width=\columnwidth]{wmt20enja-joeynmt.png}
%     \caption{wmt20 enja joeynmt}
%     \label{fig:enja}
% \end{figure}

\subsubsection{Prompt-Based Machine Translation using LLM}

To evaluate the performance of the proposed method on large language models (LLM), we decode an LLM using the proposed decoding strategy. 
We conduct experiments using BLOOMZ and mT0 model \cite{muennighoff-etal-2023-crosslingual} available in HuggingFace's Transformers library \cite{wolf-etal-2020-transformers}.\footnote{\url{https://huggingface.co/bigscience/bloomz-7b1-mt}} We load the model with 8-bit precision to reduce memory consumption.
We use the translation prompt as a guide to trigger its translation capability, which is shown to be efficient by \citet{jiao2023chatgpt}:

% \texttt{Please provide the [TGT] translation for these sentences: }

% \todo{what's the best way to force the prompt right here?}
\begin{table}[H]
    \texttt{Please provide the French translation for these sentences: }
    % \caption{Translation prompt}
    % \label{tab:prompt}
\end{table}

\begin{figure}
    \centering
    \includegraphics[width=\columnwidth]{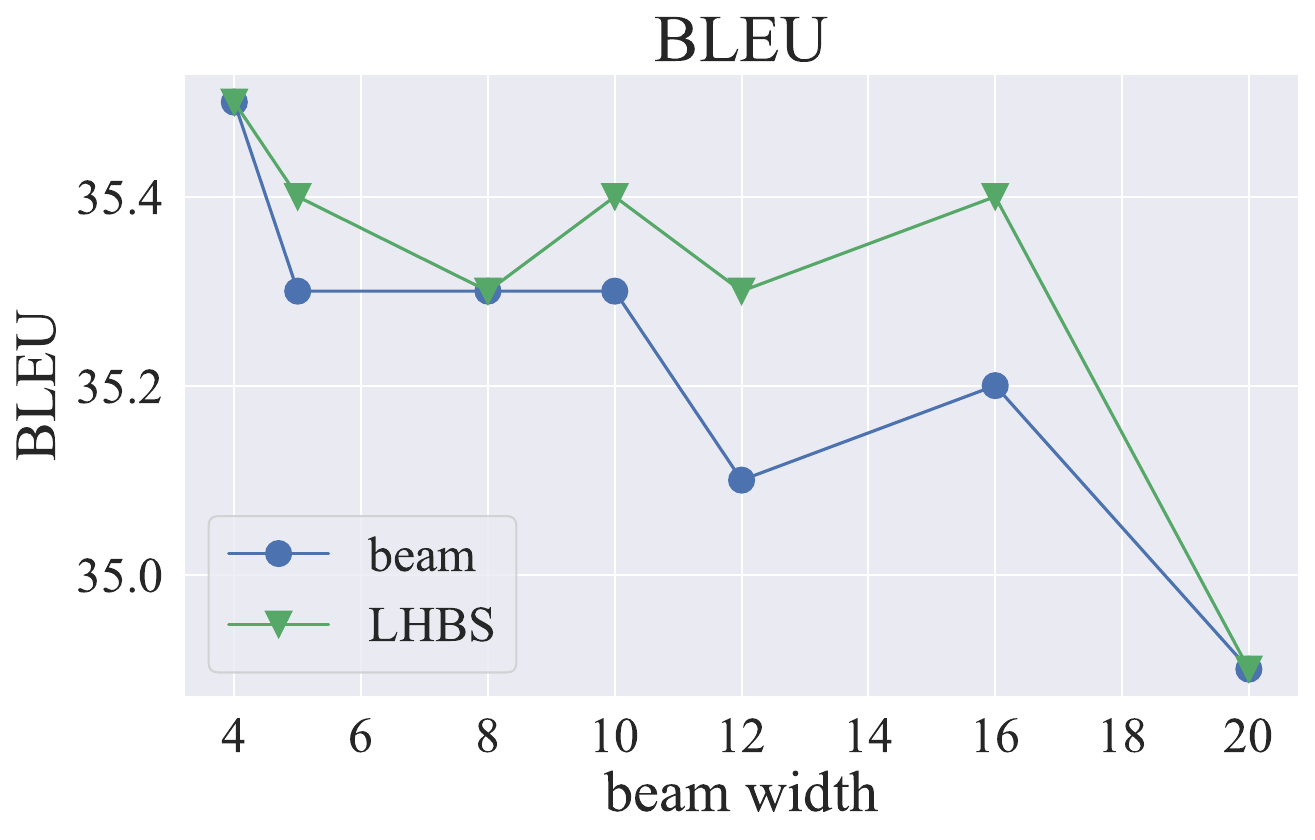}
    \caption{Results on prompt-driven machine translation with LLM (Bloomz and mT0). Evaluated on the first 200 sentences of WMT'14 En-Fr.}
    \label{fig:llm}
\end{figure}

% Note that 
Experiments are performed on the first 200 sentences of WMT'14 En-Fr dataset \cite{Bojar2014}  due to computational constraints.
% \todo{need to mention?} 
Note that because WMT'14 En-Fr test dataset is public, they are likely to be included in the training data of the LLM directly or indirectly. We believe that it does not significantly distort the result of the comparison of the decoding strategies.
Figure \ref{fig:llm} reports the BLEU score of the experiment. We observe that the proposed method is improved upon beam search.

% \begin{table}
%     \centering
%     \begin{tabular}{c|c}
%         PLACEHOLDER & PLACEHOLDER \\
%         PLACEHOLDER & PLACEHOLDER
%     \end{tabular}
%     \caption{Results on machine translation with LLM (Bloomz and mT0). BLEU score predictions generated with \Proposed{} and beam search. Best scores per beam size are bolded.}
%     \label{tab:llm}
% \end{table}

% \begin{table}
%     \centering
%     \begin{tabular}{c|c|ccc}
%     \toprule
%     Dataset & Decoder & $k=5$ & $k=10$ & $k=20$ \\
%     \hline \hline
%     \multirow{2}{*}{En-Fr}  & beam & 35.30 & 35.30 & \textbf{34.90} \\
%     & \pro{}  & \underline{\textbf{35.40}} & \underline{\textbf{35.40}} & \textbf{34.90} \\
%     \bottomrule
%     \end{tabular}
%     \caption{Results on prompt-driven machine translation with LLM (Bloomz and mT0). Evaluated on the first 200 sentences of WMT'14 En-Fr. The best scores per beam size are bolded.}
%     \label{tab:llm}
% \end{table}

\subsection{Text Summarization}

We evaluate the proposed method on two datasets, CNN/DM \cite{hermann2015teaching} and XSum \cite{narayan-etal-2018-dont}.
We use the BART-Large model fine-tuned on each dataset available from fairseq library \cite{lewis-etal-2020-bart}.\footnote{\url{https://github.com/facebookresearch/fairseq/tree/main/examples/bart}}
% We used the best parameter suggested for the pretrained models. We use the same hyperparameters for the proposed method and did not run any hyperparameter tuning. 
We use ROUGE-L \cite{lin-2004-rouge} as the evaluation metric.\footnote{\url{https://github.com/pltrdy/files2rouge}}
Table \ref{tab:sum} shows the ROUGE-L score on the two datasets.
On both datasets and across different beam widths, the results indicate that the proposed method produces higher quality summarizations.

% \todo{would it be better to run hyperparameter tuning? it can improve the performance. is it actually useful? probably we can compare the performance of the optimal parameter vs. current best. how is hyperparameter tuned in training the model? let's use the same validation set for tuning.}

\begin{table}
    \centering
    \begin{tabular}{c|c|ccc}\toprule
        Dataset & Decoder  & $k=5$    & $k=10$   & $k=20$   \\ \hline \hline
        \multirow{2}{*}{CNN/DM} & beam & 40.6 & 40.2 & 39.9 \\
        & \pro{}    & \underline{\textbf{40.7}} & \textbf{40.5} & \textbf{40.2} \\ \hline        
        \multirow{2}{*}{XSum}   & beam & 35.1 & 35.1 & 35.0 \\
           & \pro{}    & \underline{\textbf{35.3}} & \underline{\textbf{35.3}} & \textbf{35.1} \\
        \bottomrule
    \end{tabular}
    \caption{ROUGE-L score for predictions generated with \Proposed{} (\pro{}) and beam search. The best scores per beam size are bolded. The best score is underlined.}
    \label{tab:sum}
\end{table}

\section{Related Work}

The phenomenon that using a larger beam leads to worse performance has been analyzed in a number of studies \cite{koehn-knowles-2017-six,murray-chiang-2018-correcting,yang-etal-2018-breaking,stahlberg-byrne-2019-nmt,pmlr-v97-cohen19a,leblond-etal-2021-machine}.
Many of the authors observe that widening the beam search degrades performance due to a bias in sequence models to favor shorter sequences even with a length penalty.
% Interestingly, we observe that sequences tend to become longer by searching deeper.
Other authors have investigated why beam search successfully generates high quality sequences.
The uniform information density hypothesis \cite{Levy2005,Levy2006} is introduced to explain why beam search outperforms exhaustive search \cite{meister-etal-2020-beam,meister-etal-2021-revisiting}. They hypothesize that narrowing the width of beam search induces a bias in the decoding that enforces uniform information density, resulting in higher quality sequences.
Although many have studied the width of the beam search, little is known about the depth of the search. 
Our work extends the analysis to the search depth and observes a similar trade-off between search and UID error, which is balanced by the lookahead depth parameter.

Some authors have studied lookahead strategies for decoding.
\citet{hargreaves-etal-2021-incremental} investigates the greedy roll-out strategy to apply reranking during decoding instead of only at the end.
\citet{lu-etal-2022-neurologic} evaluated several lookahead strategies to estimate the future score of the given partial hypothesis.
Several works have investigated the lookahead strategy for constraint sentence generation tasks using Monte Carlo sampling \cite{miao2019cgmh,zhang-etal-2020-language-generation,leblond-etal-2021-machine}.
% \citet{leblond-etal-2021-machine} investigates value-based decoding.
% Prior work focus on sampling based lookahead while we investigate tree-based lookahead.
Our analysis provides a fundamental insight into why these lookahead strategies can be effective. Furthermore, while these methods require additional calls to the scoring function, we present a decoding strategy that does not require any additional calls. 
% \todo{In addition to the analysis we present \pro{} which is computationally feasible}

% \todo{should i mention? easy refute is "do experiments and resubmit"} \todo{Put it to Section 2?}
This work focuses on the quality of the text evaluated by its similarity to the reference text. Previous work has investigated other factors such as diversity \cite{vijayakumar2018diverse,meister-etal-2021-determinantal}, constraints \cite{anderson-etal-2017-guided,hokamp-liu-2017-lexically}, or faithfulness \cite{king-etal-2022-dont,wan-etal-2023-faithfulness}. 
How the lookahead strategy affects these factors is an open question.

% \todo{How should we mention monotonic beam search? The orignal idea came from it, but it is no longer related...} 
The idea of sequentially processing beam slots in \pro{} is inspired by the monotonic beam search \cite{Lemons2022}. While they use the technique to guarantee the monotonicity of the beam search algorithm, we use it to incorporate the score of the previous step for hypothesis selection.
% \todo{Remove this? monotonic beam search is no longer a scope of the paper.} We observe the performance of monotonic beam search for NMT is on par with beam search. 

\section{Conclusions}

% The research question of this study is whether there is a better trade-off between beam search and MAP decoding.
To study the effect of search depth on the performance of decoding strategies for text generation models, we introduce \Analyzed{} (\ana{}). % Lookahead beam search is a search algorithm that can tune its search locality by the lookahead depth. 
% Beam search and MAP decoding are a special case of \Analyzed{} with lookahead depth set to $0$ and $\infty$, respectively.
\ana{} is a generalization of beam search and exhaustive search that allows control of the search depth by its hyperparameter.
In our machine translation experiments, we observe that \ana{} outperforms beam search, suggesting that there is room for beam search to improve by searching deeper.
We observe that increasing lookahead depth reduces search error but increases UID error, similar to the observation reported by \citet{meister-etal-2020-beam} for increasing beam width. The result suggests that \ana{} is able to find a better trade-off between the two errors, thus achieving better performance.

% We proposed $t$-step surprisal, a generalization of the surprisal to consider multiple steps of tokens for quantifying the surprisal. Using the concept of $t$-step surprisal, we proposed a semi-local regularization to operationize the uniform information density.
Although promising, \ana{} is markedly computationally intensive.
We present \Proposed{} (\pro{}), a variant of beam search that heuristically simulates \ana{} with a 1-step lookahead in the same computational time as beam search,
\pro{} is simple, domain-independent, and requires no additional training.
In our empirical evaluation, we find that \pro{} improves over vanilla beam search on machine translation and text summarization tasks overall.

\section*{Acknowledgments}
We extend our sincere gratitude to all those who offered their perspectives on this research. We are especially grateful to Sho Hoshino and Masato Mita for their insightful comments.

% \bibliographystyle{aaai22}
% \bibliography{export}

\appendix
\counterwithin{figure}{section}
\counterwithin{table}{section}
\counterwithin{listing}{section}
% \section{\analyzed{}}

% \section{Implementation of \analyzed{}}

% \todo{do we need to?} 
% We implement the lookahead strategy using best-first branch-and-bound.

\section{Implementation of \analyzed{}}

Algorithm \ref{lst:analyzed} describes the procedure of \analyzed{}. 
To reduce the computation time, we implement the evaluation of $h_d$ by best-first branch-and-bound search (Algorithm \ref{lst:analyzed} and \ref{lst:eval}). We choose best-first branch-and-bound over Dijkstra because it requires less memory consumption ($O(d + |\vocab|)$). 
Since the scoring function is monotonically decreasing \cite{meister-etal-2020-best}, we can prune a partial hypothesis that is lower than the current $k$-th largest score before expanding the hypothesis further.
The function $\kthmax_{\vy \in Y'}(f(\vy))$ returns the $k$-th largest score among $Y'$ if $|Y'| \geq k$ and negative infinity otherwise.
We explore the candidates in best-first order -- the hypothesis with the highest score is explored first. In this way, we have a higher chance of pruning the less promising hypothesis, thus reducing computation.
Note that the algorithm is guaranteed to find the same path as breadth-first search, thus preserving the result of the decoding.

\begin{listing}[p]
\caption{\Analyzed{}}
\label{lst:analyzed}
\renewcommand{\algorithmicrequire}{\textbf{Input:}}
\renewcommand{\algorithmicensure}{\textbf{Output:}}
\begin{algorithmic}[1]
\REQUIRE a set of hypotheses $Y_{t-1}$ of length $t-1$ \\
\ENSURE a set of hypotheses $Y_t$ of length $t$
\STATE $\mathcal{B} = \{\vy_{t-1} \circ y | y \in \mathcal{\bar{V}}\}$ \\
\STATE $\{\vy_{t}^1, \vy_{t}^2,...,\vy_{t}^b\} = \text{sort}(\mathcal{B})$ in a descending order of $p(\vy_{t}^i | \vx)$
\STATE $Y' \gets \emptyset$
\STATE $b \gets |\bar{\vocab}|$
\FOR {$i \in \{1,...,b\}$}
    \IF {$\log p_\theta(\vy_{t}^i | \vx) < \kthmax_{\vy \in Y'}(f(\vy))$}
        \RETURN $Y_t = \argtopk_{\vy \in Y'}(f(\vy))$
    \ENDIF
    \STATE $f(\vy_{t}^i) \gets Eval(\vy_t^i, d, \kthmax_{\vy \in Y'}(f(\vy)))$
    \IF {$f(\vy_{t}^i) > \kthmax_{\vy \in Y'}(f(\vy))$}
        \STATE $Y' \gets Y' \cup \{\vy_t^i\}$
    \ENDIF
\ENDFOR
\RETURN $Y_t = \argtopk_{\vy \in Y'}(f(\vy))$
\end{algorithmic}
\end{listing}

\begin{listing}[p]
\caption{$Eval(\vy_t, d, f_{\mathrm{max}})$}
\label{lst:eval}
\renewcommand{\algorithmicrequire}{\textbf{Input:}}
\renewcommand{\algorithmicensure}{\textbf{Output:}}
\begin{algorithmic}[1]
\REQUIRE a hypothesis $\vy_t$, a depth $d$, and a threshold $f_{\mathrm{max}}$ \\
\ENSURE a score of the hypothesis $h_d(\vy)$
\IF {$d = 0$}
    \RETURN $\log p_\theta(\vy_t | \vx)$
\ENDIF
\STATE $\mathcal{B} = \{\vy_t \circ y | y \in \mathcal{\bar{V}}\}$
\STATE $\{\vy_{t+1}^1, \vy_{t+1}^2,...,\vy_{t+1}^b\} = \text{sort}(\mathcal{B})$ in a descending order of $\log p_\theta(\vy_{t+1}^i | \vx)$
\FOR {$i \in \{1..b\}$}
    \IF {$\log p_\theta(\vy_{t+1}^i) < f_{\mathrm{max}}$}
        \RETURN $f_{\mathrm{max}}$
    \ENDIF
    \STATE $f_i \gets Eval(\vy_{t+1}^i, d-1, f_{\mathrm{max}})$
    \IF {$f_i > f_{\mathrm{max}}$}
        \STATE $f_{\mathrm{max}} \gets f_i$
    \ENDIF
\ENDFOR
\RETURN $f_{\mathrm{max}}$
\end{algorithmic}
\end{listing}

% \clearpage

% \section{Analysis of Search Depth}

% \subsection{BLEU Score}

% Table \ref{tab:bleu} reports the BLEU score on the first 100 sentences of WMT'14 En-Fr and WMT'14 En-De. Overall, lookahead depth of 2 is most frequently the best for each beam width with lookahead depth of 1 and 3 occasionally being the best. 

% \subsection{UID Error}
% % The question is if searching deeper has an effect on UID, and thus on BLEU score.
% % The standard deviation of surprisals is reported as a function of beam width (Figure  \ref{fig:search-uid-width}) and lookahead depth (Figure \ref{fig:search-uid-depth}).
% % We observe that both beam width and lookahead depth have a negative correlation to UID.

% Figure \ref{fig:uid} shows the standard deviation of the surprisal and BLEU with varying numbers of lookahead depth and beam width. 
% We do not observe a strong correlation between them. 
% Although \ana{} with 1- and 2-step lookaheads have higher BLEU scores than beam search, they also have higher average standard deviation per sentence of surprisals. Therefore, UID error alone cannot account for the performance of \ana{}.

% \begin{figure}
%     \centering
%     \includegraphics[width=\columnwidth]{depth-ustddev.pdf}
%     \caption{Average standard deviation of surprisals per sequence and BLEU for translations generated using lookahead beam search with varying beam widths and lookahead steps.}
%     \label{fig:uid}
% \end{figure}

\section{Evaluation of Exhaustive Search (MAP Decoding)}

We perform an exhaustive search on the first 100 sentences of WMT'14 En-Fr and WMT'14 En-De datasets. The results are summarized in Table \ref{tab:map}.
Overall we observe a significant decrease in BLEU and sequence length, as observed in previous work \cite{stahlberg-byrne-2019-nmt}.

\begin{table}
    \centering
    \begin{tabular}{c|cc}
    \toprule
    Dataset & En-Fr & En-De \\ \hline \hline
    BLEU & 2.2 & 6.0 \\
    sequence length & 9.169 & 16.217 \\
    negative log-likelihood & 8.195 & 8.246 \\
    stddev of surprisals & 0.291 & 0.486 \\
    \bottomrule
    \end{tabular}
    \caption{Results of exhaustive search (MAP decoding) on the first 100 sentences of WMT'14 En-Fr and En-De datasets.}
    \label{tab:map}
\end{table}

\section{Additional evaluations of \analyzed{}}

\subsubsection{Wall-Clock Time}

\begin{figure}
    \centering
    \includegraphics[width=\columnwidth]{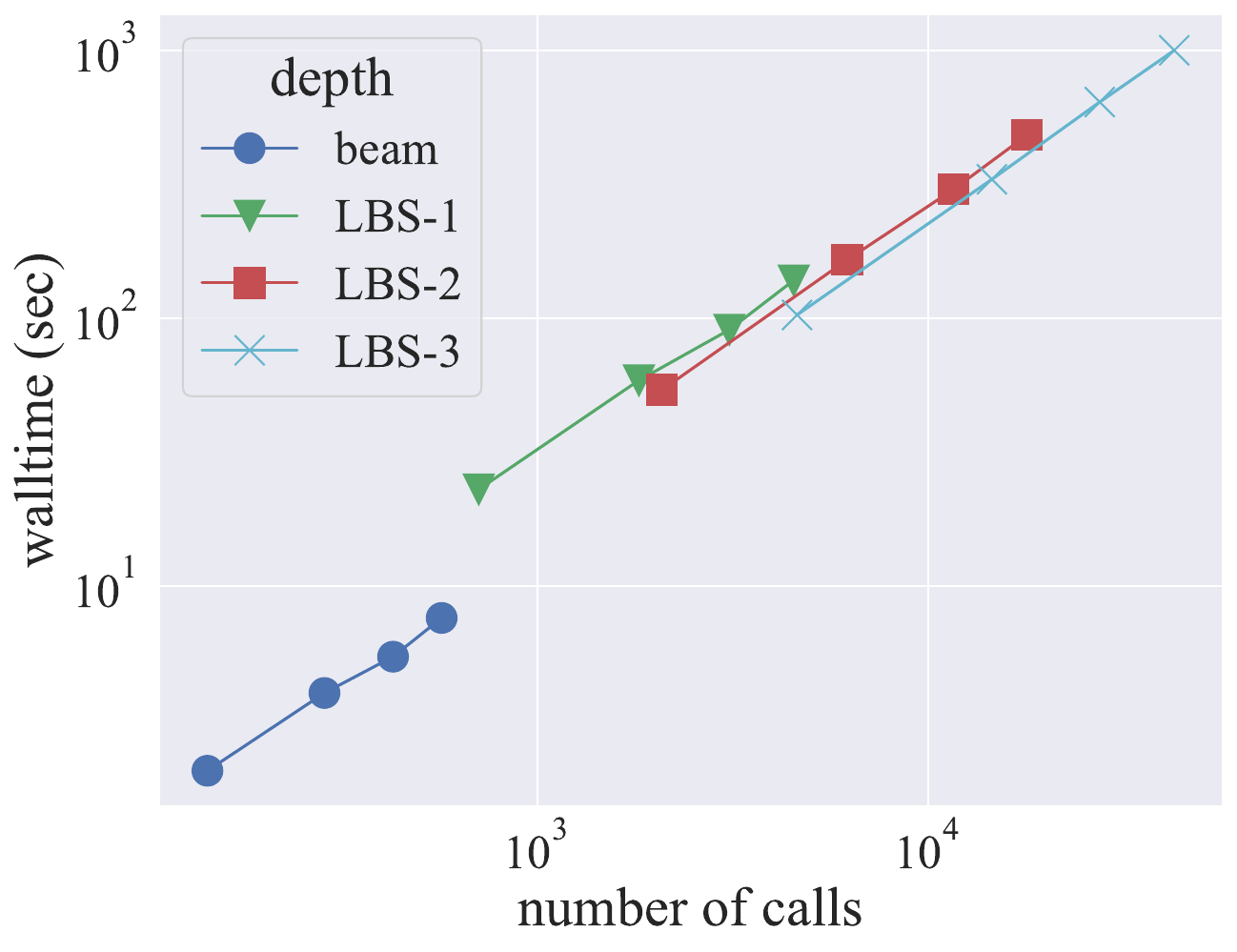}
    \caption{Comparison of the average number of calls to the scoring function to the wall-clock time (WMT'14 En-Fr).}
    \label{fig:expansions-wtime}
\end{figure}

Table \ref{tab:nexpansions} reports the number of calls to the scoring function (e.g. probabilistic model) by \analyzed{}.
We observe that the number of calls grows rapidly with increasing lookahead depth. The wall-clock time is mostly proportional to the number of calls (Figure \ref{fig:expansions-wtime}).
Note that the wall-clock time depends on the hardware.
All the experiments are performed on \texttt{g4dn.xlarge} instances on AWS EC2 (4 vCPU cores, 16 GB memory, and an NVIDIA T4 GPU).

\begin{table}[p]
    \centering
    % \begin{tabular}{l|rrrr}
    % \toprule
    % Decoder & $k=1$ & $k=5$ & $k=10$ & $k=15$  \\
    % \hline \hline
    % beam & 29.06 & 143.38 & 285.63 & 427.77 \\
    % \ana{}-$1$ & 60.86 & 708.70 & 1820.14 & 3099.95 \\
    % \ana{}-$2$ & 98.70 & 2084.45 & 6209.03 & 11616.00 \\
    % \ana{}-$3$ & 146.92 & 4617.61 & 14544.80 & 27445.60 \\
    % \bottomrule
    % \end{tabular}
    \begin{tabular}{c|rrrr}
    \toprule
    Decoder & $k=5$ & $k=10$ & $k=15$ & $k=20$  \\
    \hline \hline
    beam & 145.86 & 291.38 & 436.07 & 580.99 \\
    \ana{}-1 & 718.02 & 1841.83 & 3142.28 & 4590.62 \\
    \ana{}-2 & 2103.73 & 6270.25 & 11630.90 & 17946.10 \\
    \ana{}-3 & 4656.60 & 14471.00 & 27364.80 & 42597.90 \\
    \bottomrule
    \end{tabular}
    \caption{Average number of calls to the scoring function (probabilistic model) per sentence (WMT'14 En-Fr).}
    \label{tab:nexpansions}
\end{table}

% \clearpage

% \subsection{Additional evaluations of \analyzed{}}

\subsubsection{Results on WMT'14 En-De}

% We report the results of the \analyzed{} on WMT'14 En-De dataset.
% Overall, we observe qualitatively similar results to the experiments on WMT'14 En-Fr dataset.
% Figure \ref{fig:locality2} 
Table \ref{tab:bleu} reports the BLEU score on the first 100 sentences of WMT'14 En-De dataset. Overall, we observe that the BLEU score increases with the lookahead strategy.
The highest BLEU score is achieved when using a beam width of 5 and a lookahead depth of 1 and 2.

% The perplexity is reported in Figure \ref{fig:perlexity2}. The average sequence length is reported in Table \ref{tab:length2}. The correlation of beam width and lookahead depth are $-0.81$ and $0.51$, respectively.

\begin{table}[p]
\centering
\begin{tabular}{c|c|rrrr}
\toprule
Dataset & Decoder & $k=5$ & $k=10$ & $k=15$ & $k=20$ \\
\hline \hline
% \multirow{4}{*}{En-Fr (400)} 
%     & beam & 35.5 & 35.5 & 35.4 & 35.4 \\
% & \ana{}-1 & 35.7 & 35.8 & 35.7 & \textbf{35.6} \\
% & \ana{}-2 & \underline{\textbf{35.9}} & \underline{\textbf{35.9}} & \textbf{35.8} & \textbf{35.6} \\
% & \ana{}-3 & 35.7 & 35.5 & 35.7 & 35.5 \\
% \multirow{4}{*}{En-Fr} & beam & 35.3 & 35.5 & 35.4 & 35.2 \\
% & \ana{}-$1$  & 35.8 & 35.7 & 35.6 & 35.5 \\
% & \ana{}-$2$ & \underline{\textbf{36.1}} & \underline{\textbf{36.1}} & \textbf{35.7} & \textbf{35.7} \\
% & \ana{}-$3$ & 35.7 & 35.9 & 35.6 & 35.4 \\ \midrule
% \midrule
\multirow{4}{*}{En-De} 
      & beam & 22.7 & 21.9 & 22.0 & 21.8 \\
& \ana{}-$1$ & \underline{\textbf{23.2}} & 22.6 & 22.2 & 21.5 \\
& \ana{}-$2$ & \underline{\textbf{23.2}} & \textbf{22.7} & 22.6 & 22.6 \\
& \ana{}-$3$ & 23.0 & \textbf{22.7} & \textbf{23.0} & \textbf{23.0} \\
\bottomrule
\end{tabular}
\caption{Evaluation of \analyzed{} on the first 100 sentences of WMT'14 En-De datasets. The best for each beam width is bolded. The best score is underlined.}
\label{tab:bleu}
\end{table}

\subsubsection{Fully Convolutional Decoder}

To test the effect of the lookahead strategy on a non-Transformer model, we evaluate the performance of \ana{} on a fully convolutional decoder proposed by \citet{pmlr-v70-gehring17a}.
For reproducibility, we use the pretrained model provided by fairseq.\footnote{\url{https://github.com/facebookresearch/fairseq/tree/main/examples/translation}}
Table \ref{tab:conv} reports the BLEU score. We observe that \ana{} outperforms beam search overall.

\begin{table}[p]
    \centering
    \begin{tabular}{c|c|rrrr}
    \toprule
    Dataset & Decoder & $k=5$ & $k=10$ & $k=15$ & $k=20$ \\
    \hline \hline 
    \multirow{4}{*}{En-Fr} & beam & 35.3 & 35.5 & 35.4 & 35.2 \\
    & \ana{}-1 & 35.8 & 35.7 & 35.6 & 35.5 \\
    & \ana{}-2 & \underline{\textbf{36.1}} & \underline{\textbf{36.1}} & \textbf{35.7} & \textbf{35.7} \\
    & \ana{}-3 & 35.7 & 35.9 & 35.6 & 35.4 \\
    \bottomrule
    \end{tabular}
    \caption{BLEU on the first 100 sentences of WMT'14 En-Fr using a fully convolutional decoder. The best for each beam width is bolded. The best score is underlined.}
    \label{tab:conv}
\end{table}

\subsubsection{Evaluation of \ana{}-1 on the entire WMT'14 En-Fr}

To evaluate the lookahead strategy on larger samples, we evaluate \ana{}-1 on the entire WMT'14 En-Fr dataset.
Due to computational constraints, we present only the evaluation of \ana{}-1. Table \ref{tab:entire} reports the BLEU score. We observe that \ana{}-1 achieves higher BLEU compared to beam search (except for En-Fr with $k=15$).

\begin{table}[p]
    \centering
    \begin{tabular}{c|c|rrrr}
    \toprule
    Dataset & Decoder & $k=1$ & $k=5$ & $k=10$ & $k=15$ \\
    \hline \hline 
    \multirow{2}{*}{En-Fr} & beam & 34.8 & 35.8 & 36.0 & \textbf{36.0} \\
    & \ana{}-1 & \textbf{35.2} & \textbf{35.9} & \underline{\textbf{36.1}} & 35.9 \\\hline
    \multirow{2}{*}{En-De} & beam & 28.6 & 29.3 & 29.0 & 28.9 \\
    & \ana{}-1& \textbf{28.8} & \underline{\textbf{29.4}} & \textbf{29.2} & \textbf{29.0} \\
    \bottomrule
    \end{tabular}
    \caption{BLEU on the entire dataset on WMT'14 En-Fr and En-De. The best for each beam width is bolded. The best for each dataset is underlined.}
    \label{tab:entire}
\end{table}

\subsubsection{Results on the last 50 Sentences of WMT'14 En-Fr}

We perform an evaluation on the last 50 sentences of WMT'14 En-Fr. Table \ref{tab:50} reports the BLEU score. We observe that \ana{} improves over beam search overall.

\begin{table}
    \centering
    \begin{tabular}{c|rrrr}
    \toprule
    Decoder & $k=1$ & $k=5$ & $k=10$ & $k=15$ \\
    \hline \hline 
        beam & \textbf{37.6} & 37.2 & 35.9 & 35.6 \\
    \ana{}-1 & 37.3 & 37.4 & \textbf{36.4} & \textbf{36.5} \\
    \ana{}-2 & 36.9 & \underline{\textbf{38.1}} & 36.2 & 35.9 \\
    \ana{}-3 & 36.6 & 37.4 & \textbf{36.4} & 36.4 \\
    \bottomrule
    \end{tabular}
    \caption{BLEU on the last 50 sentences of WMT'14 En-Fr. The best for each beam width is bolded. The best score is underlined.}
    \label{tab:50}
\end{table}

% \clearpage

\section{Additional evaluation of \proposed{}}

% \subsection{Results on Joey NMT}

% There are several variations in beam search implementation. To test the performance of \proposed{} on other beam search implementation, we implemented the method on Joey NMT library.
% One of the design choice is whether to generate $2k$ candidates instead of $k$ candidates to take into account of EOS strings. fairseq and huggingface library take this strategy. Joey NMT does not. 

To test the performance of \proposed{} on other NMT systems, we perform an evaluation on Joey NMT library \cite{kreutzer-etal-2019-joey}. The experiments are performed on WMT'14 En-De dataset, since the pretrained model on the task is provided by Joey NMT library.
Table \ref{tab:translate-joey} shows the BLEU scores. Overall, we observe that the proposed method improves over beam search.

\begin{table}[h]
    \centering
    \begin{tabular}{c|rrrr}
    \toprule
    Decoder & $k=4$ & $k=8$ & $k=16$ & $k=32$ \\
    \hline \hline
    beam & 26.06 & \textbf{26.36} & 26.20 & 25.77 \\
    \pro{} & \textbf{26.20} & 26.20 & \underline{\textbf{26.38}} & \textbf{25.90} \\
    \bottomrule
    \end{tabular}
    \caption{BLEU score on WMT'14 En-De using Joey NMT library. Best per beam width is bolded.}
    \label{tab:translate-joey}
\end{table}

% \subsection{Experiments on Fully Convolutional Models}

% \subsection{Uniform Information Density}
% We use SGNMT library to decode the models.

% \section{Ethical Considerations}

% While language generation can be used for malicious purposes, we do not foresee any specific ethical concerns with the analysis in this paper beyond those discussed in \citet{Bender2021}.

% \bibliography{export}

\end{document}